\documentclass[10pt,twocolumn,letterpaper]{article}
\usepackage[utf8]{inputenc}
\usepackage{eda_preprint}
\usepackage[T1]{fontenc}
\usepackage{refcount}

\usepackage{placeins}
\newsavebox{\paperTableBox}

\usepackage{tcolorbox}
\tcbuselibrary{breakable,skins}
\newtcolorbox{agentprompt}[1]{enhanced,breakable,
 colback=white,colframe=black!62,colbacktitle=black!62,coltitle=white,
 fonttitle=\bfseries,fontupper=\small,title={#1},
 title after break={#1 (continued)},
 boxrule=0.5pt,arc=2mm,left=8pt,right=8pt,top=7pt,bottom=7pt,
 before skip=8pt,after skip=8pt,pad at break*=3mm}

\newtcolorbox{agentpromptlong}[1]{enhanced,breakable,
 colback=white,colframe=black!62,colbacktitle=black!62,coltitle=white,
 fonttitle=\bfseries,fontupper=\small,title={#1},
 title after break={#1 (continued)},
 boxrule=0.5pt,arc=2mm,left=8pt,right=8pt,top=7pt,bottom=7pt,
 before skip=8pt,after skip=8pt,pad at break*=3mm}

\usepackage{dblfloatfix}

\definecolor{cvprblue}{rgb}{0.21,0.49,0.74}
\usepackage[breaklinks,colorlinks,allcolors=cvprblue]{hyperref}
\usepackage[capitalize]{cleveref}
\title{EDA: Observe, Verify, and Assess for\\ Evidence-Grounded Multimodal Depression Analysis}
\author{\begin{tabular}{c}
Changzeng Fu\textsuperscript{1}, Ming Shang\textsuperscript{2}, Xinyi Zhang\textsuperscript{2},\\[3pt]
Jiadong Wang\textsuperscript{3}, Runhao Zeng\textsuperscript{4}, Shiqi Zhao\textsuperscript{1}\\[6pt]
\small\textsuperscript{1}Northeastern University\quad\textsuperscript{2}Northeastern University at Qinhuangdao\\[2pt]
\small\textsuperscript{3}Technical University of Munich\quad\textsuperscript{4}Shenzhen MSU-BIT University\\[4pt]
\small\texttt{fuchangzeng@qhd.neu.edu.cn}
\end{tabular}}
\date{}
\hypersetup{pdftitle={EDA: Observe, Verify, and Assess for Evidence-Grounded Multimodal Depression Analysis},pdfauthor={Changzeng Fu, Ming Shang, Xinyi Zhang, Jiadong Wang, Runhao Zeng, Shiqi Zhao},pdfsubject={Multimodal depression analysis},pdfkeywords={depression assessment,evidence grounding,multi-agent reasoning}}
\begin{document}
\twocolumn[\maketitle
\begin{minipage}{\textwidth}
\begin{abstract}
Existing multimodal depression assessment methods often summarize long recordings into global behavioral descriptions. This leaves \textbf{the source and decision relevance of individual observations difficult to inspect}. We propose \textbf{EDA}, an evidence-grounded framework organized around \textbf{Observe, Verify, and Assess}. In \textbf{Observe}, literature-informed acoustic and visual cues are localized into source-linked Evidence Records with temporal intervals and behavioral measurements. In \textbf{Verify}, frozen multimodal agents examine whether the cited media support each observation. The simplified Support admission policy retains positively verified records and their review grounds. In \textbf{Assess}, statistical descriptors and temporal events support reference-based behavioral comparison, while PHQ-9-guided text analysis provides complementary symptom evidence. Cross-modal disagreement triggers review of the original text before final selection. Experiments on D-Vlog and LMVD evaluate human-referenced cue agreement and downstream assessment utility separately. AI cue annotations achieve 91.02\% accuracy against human consensus labels on shared source segments. With a frozen 7B backbone, EDA Support improves F1 over direct prediction by 7.61 and 9.88 percentage points on D-Vlog Leave-Query-Out and LMVD, respectively. These findings support explicit, source-linked behavioral evidence for inspectable multimodal depression assessment.
\end{abstract}

\end{minipage}\vspace{10pt}]
\raggedbottom
\begin{figure}[!t]
    \centering
    \includegraphics[width=\linewidth]{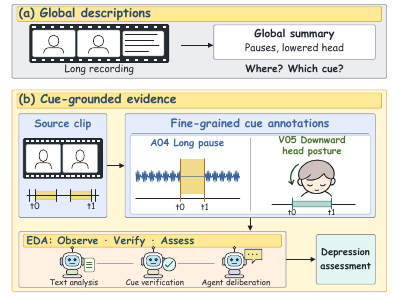}
    \caption{\textbf{Motivation.} From global descriptions to cue-grounded multi-agent assessment.}
    \label{fig:motivation}
\end{figure}

\section{Introduction}
\label{sec:introduction}

Video-based depression assessment uses both spoken content and observable behavior. Transcripts can describe symptoms and personal experiences. Vocal expression, facial activity, and head movement provide additional information~\cite{mundt2007voice,girard2014withdrawal,aldeneh2019mood}. In long recordings such as D-Vlog and LMVD~\cite{yoon2022dvlog,he2026lmvd}, these behaviors vary over time and across contexts. A useful assessment must identify relevant cues, locate their source, and explain how they inform the final judgment.

Current methods address different parts of this task. Fusion models learn acoustic--visual interactions~\cite{tao2024stst,ye2024depmamba}, while symptom-oriented language models organize transcript passages by depression-related domains~\cite{zhang-etal-2025-explainable,zheng2026explainmdrc}. DepressionAgent adds modality-specific extraction, support--challenge deliberation, and cross-modal reflection~\cite{zhu2026depressionagent}. Yet a recording-level evidence explanation still condenses a long video into a short narrative (Fig.~\ref{fig:motivation}a). It tells readers \emph{what} the model observes. A statement such as \emph{frequent pauses} names a behavior, but does not identify its source intervals. Readers still need to know \emph{where/when} it occurs and how it affects the assessment. This leads to three linked questions. \textbf{Observe: What behavioral evidence is observable?} Acoustic and visual behaviors need reproducible definitions and source intervals. \textbf{Verify: Is the observation supported by its source?} The cited media and measurements must support each candidate before it is retained. \textbf{Assess: How should verified evidence inform assessment?} Behavioral comparison, symptom-related text, and conflict review must have explicit roles in the final judgment.

We introduce \textbf{EDA} around \textbf{Observe, Verify, and Assess}. Observe operationalizes five acoustic and four visual cues from prior literature. Their annotations use no depression labels and retain source intervals and measurements. Verify checks these records against the original media. Its Evidence mechanism combines Support verification, Counter challenge, and cue-level adjudication, retaining the grounds for admission or exclusion. The optimized Support policy admits positively verified records directly and forms our final model. Assess compares retained behavioral records with labeled references, using statistical descriptors and temporal events. In parallel, PHQ-9-guided text agents examine symptom candidates and their context. Cross-modal disagreement triggers textual source review and an external acceptance gate. Thus, EDA connects each observation to its source, verification, and subsequent use.

Our experiments examine each step. Cue distributions show that the same behavior has different associations with depression across datasets. Cue interpretation therefore depends on the dataset. AI cue annotations achieve 91.02\% accuracy against human consensus labels, pooled across D-Vlog and LMVD. This check evaluates cue agreement on shared source segments. For final assessment, 7B EDA Support improves F1 over direct prediction by 7.61 points in the D-Vlog reference setting and 9.88 on LMVD. In the corresponding D-Vlog comparison with a larger-model baseline, accuracy and F1 improve by 1.87 and 1.30 points. Admission comparisons support the optimized Support mechanism. Representation controls show that events need not improve classification, although their source-linked timing remains inspectable. Evidence quality and assessment utility are therefore evaluated separately. Our contributions are:
\begin{itemize}
    \item \textbf{Observe: construct source-linked behavioral evidence.} We operationalize five acoustic and four visual cues with literature-informed, label-independent criteria. Evidence Records retain each observation's temporal anchors and measurements.
    \item \textbf{Verify: ground evidence admission in its source.} We develop an explicit verification mechanism with source-based review grounds. The simplified Support policy optimizes admission while preserving this traceability.
    \item \textbf{Assess: use evidence in multimodal assessment.} We combine behavioral reference comparison, PHQ-9-guided text analysis, and disagreement-triggered review. Cue agreement, admission behavior, and assessment utility are evaluated separately.
\end{itemize}

\section{Related Work}
\label{sec:related-work}

\paragraph{Multimodal and temporal depression modeling.}
Speech and visible behavior inform depression assessment~\cite{mundt2007voice,girard2014withdrawal,aldeneh2019mood}. Multimodal models learn feature interactions across unaligned sequences~\cite{zadeh2017tfn,tsai2019mult}. Depression-specific methods study temporal fusion, instance selection, and multiscale representations~\cite{zhou2023tamfn,zhou2023caiinet,tao2024depmstat,tao2024stst,ling2024mdavif,shangguan2024attentionmil,he2024depressformer}. Recent work adds state-space modeling and behavioral interactions~\cite{ye2024depmamba,zhou2025cmfmamba,zhou2025stcmmamba,shi2026msinet,ma2026bhdd}. EDA instead makes the underlying cue records and their source support explicit.

\paragraph{Symptom evidence and agent deliberation.}
PHQ-9 defines symptom domains~\cite{kroenke2001phq}. Question-wise Modality Fusion, RED, and Explain-MDRC organize interview content around symptoms and supporting passages~\cite{mandal-etal-2025-enhancing,zhang-etal-2025-explainable,zheng2026explainmdrc}. Dep-LLM and AI Psychiatrist Assistant use evidence and specialized reasoning roles for assessment~\cite{lyu2026depllm,greene2026psychiatrist}. Multimodal language-model studies extend this analysis to speech, vision, and text~\cite{zhao-etal-2026-hears,zhang2026mllmdr}. DepressionAgent is closely related. It combines modality-specific extraction, support--challenge deliberation, arbitration, and reflection~\cite{zhu2026depressionagent}. EDA develops source-linked cue records and optimizes their admission through Support.

\paragraph{Perception, grounding, and source verification.}
VideoAgent, VideoTree, and DrVideo organize long-video content for language-model reasoning~\cite{wang2024videoagent,wang2025videotree,Ma_2025_CVPR}. VideoGLaMM studies fine-grained visual grounding~\cite{munasinghe2025videoglamm}. OPPO distinguishes cue use from input-supported emotion reasoning and uses reinforcement learning to improve perception~\cite{han2026oppo}. EDA verifies source-linked cue records with frozen models and uses Support to optimize admission. Debate and verification methods also examine alternative interpretations~\cite{du2024debate,shi2025aotd}. Its record structure separates source verification from later decision use.

\begin{figure*}[t]
    \centering
    \includegraphics[width=\textwidth]{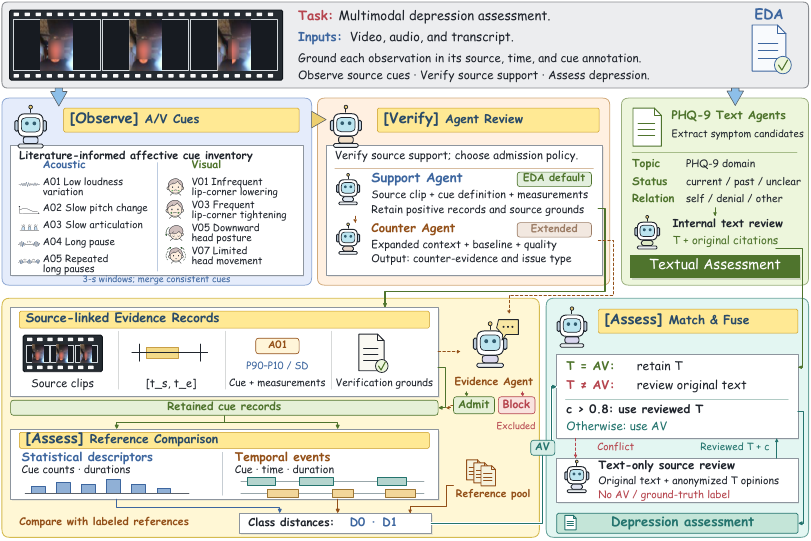}
    \caption{\textbf{EDA: Observe, Verify, and Assess.} Evidence Records connect source-linked cues, agent verification, and multimodal depression assessment.}
    \label{fig:framework}
\end{figure*}

\section{Observe: Fine-Grained Behavioral Cue Annotation}
\label{sec:cue-annotation}

EDA defines which behaviors the audio-visual workflow should observe. We derive cue concepts from the literature, specify observable criteria, and annotate their time intervals. These annotations describe behavior independently of the final depression label.

\subsection{Literature-informed Cue Definitions}
\label{sec:cue-definitions}

Prior behavioral studies guide the cue inventory. Speech studies inform low loudness variation~\cite{gumus2023daily}, slow pitch change~\cite{nilsonne1988rate,nilsonne1988speech}, slow articulation~\cite{cummins2023multilingual}, long pauses~\cite{yamamoto2020timing,mundt2007voice}, and repeated long pauses~\cite{aldeneh2019mood}. Visual-behavior studies inform infrequent lip-corner lowering, frequent lip-corner tightening, and limited head movement~\cite{girard2014withdrawal}, as well as downward head posture~\cite{fiquer2018nonverbal}. These studies inform five acoustic and four visual cue definitions (Fig.~\ref{fig:framework}). We retain the original nonconsecutive visual identifiers for record consistency.

The same definitions apply to both depression-label groups. Annotators identify the cue and its temporal extent without using the recording's depression label. Labels enter only the later reference comparison and distribution analysis. This separates the annotation of a behavior from its interpretation in depression assessment. The supplement lists the full criteria and their literature basis.

\subsection{Operationalization and Annotation Records}
\label{sec:annotation-protocol}

We construct fine-grained behavioral cue annotations through automatic interval localization and source-media review. OpenFace extracts facial action units and head pose, while FSMN-VAD identifies speech activity intervals. Cue selection uses pre-segmented 3-second audio and video windows with cue-specific measurement criteria. Adjacent windows with consistent cue assignments are merged. Source review checks the corresponding media, surrounding context, and within-recording baselines. Each record contains the recording identifier, cue label, onset, offset, source excerpt, and measurements. This Evidence Record links the source interval, cue assignment, measurements, verification grounds, and subsequent use. From these records, we construct statistical descriptors and temporal events to characterize aggregate behavioral patterns and local occurrences, respectively. The admission policies in Sec.~\ref{sec:auditing} determine which records enter behavioral comparison and retain the reasons for inclusion or exclusion. Missing media, failed tracking, and unavailable review are recorded separately from cue absence, distinguishing unavailable evidence from an unobserved behavior.

\subsection{Retained Cue Coverage by Dataset and Depression Label}
\label{sec:annotation-statistics}
\label{sec:cue-distributions}

\begin{figure}[t]
\centering
\includegraphics[width=\linewidth]{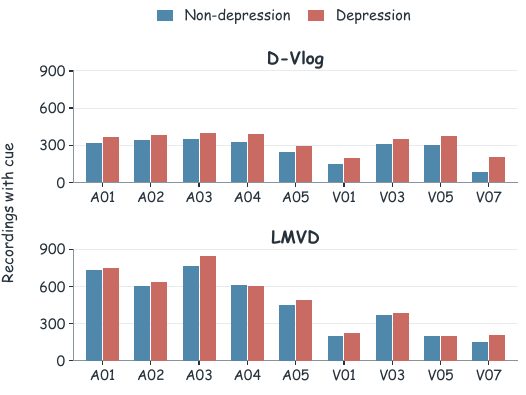}
\caption{\textbf{Numbers of depression-related cue annotations}}
\label{fig:cue-counts}
\end{figure}

Figure~\ref{fig:cue-counts} counts recordings with at least one retained annotation for each cue after Evidence adjudication. The bars separate datasets and depression-label groups. Each recording contributes at most once per cue, and cues can co-occur. These are recording counts, not interval totals or cue-detection recall. Depression labels serve only to group the resulting records in this analysis. The supplement reports media durations and admission totals.

\section{EDA: Observe, Verify, and Assess}
\label{sec:method}

\subsection{Overview and Evidence States}
\label{sec:overview}

EDA organizes source-linked cue annotation, source verification, and assessment as Observe, Verify, and Assess (Fig.~\ref{fig:framework}). It uses frozen models for source review and prediction. For a recording $X_q=(X_q^T,X_q^A,X_q^V)$, it predicts a binary depression label $\hat y_q$. Text agents analyze symptom candidates under PHQ-9 domains. In parallel, audio-visual agents check cue annotations and compare retained records with labeled references. EDA then matches the two assessments. Disagreement triggers textual review and an external acceptance gate. The Evidence mechanism defines the source-linked record and its verification process. Its extended review adds contextual challenge and cue-level adjudication. We further optimize \emph{evidence admission} through Support, which retains positively verified records directly. EDA Support combines this policy with the same reference comparison and text--AV coordination and is our final model. The framework reuses the originally blurred D-Vlog clip from InertiaMPDD. Its cue and review examples are schematic.

The evidence package retains original text passages, admitted behavioral events, and review grounds. For a candidate behavioral event,
\begin{equation}
h_i=\big(m_i,c_i,\tau_i,X_{q,\tau_i}^{m_i},\mathbf F_i\big),
\quad \tau_i=[t_i^s,t_i^e],
\label{eq:event}
\end{equation}
$m_i\in\{A,V\}$ identifies the modality, and $c_i$ identifies the cue. $X_{q,\tau_i}^{m_i}$ is the source excerpt, and $\mathbf F_i$ contains descriptors derived from the annotation. A \emph{candidate record} proposes a cue at a specific interval. A \emph{source-reviewed record} adds review judgments and their grounds. \emph{Admitted evidence} contains records selected for comparison. These states show how an observation enters assessment. They do not establish that the observation is correct or relevant.

\subsection{PHQ-9-guided Textual Evidence Analysis}
\label{sec:construction}
Text agents first retrieve symptom candidates from the original transcript using the nine PHQ-9 domains~\cite{kroenke2001phq}. Each passage records its symptom topic, temporal status, and semantic relation. These fields distinguish current personal experience from history, denial, third-party statements, and general discussion. For each domain, agents collect supporting passages, counter-evidence, and qualifying context. The assessment role considers these records alongside personal-status information and the original transcript. An internal text review checks the initial judgment. The workflow returns $\hat y_T$, a rationale, and citations to original passages. It interprets symptom evidence without assigning questionnaire scores.

The audio-visual workflow runs in parallel. It receives the cue annotations from Sec.~\ref{sec:cue-annotation} and checks each candidate against the original audio or video. It then computes a separate behavioral assessment. Neither cue annotation nor source review receives the recording's depression label.

\subsection{Source Verification and Evidence Admission}
\label{sec:auditing}

Support, Counter, and Evidence use the selected frozen multimodal model. Each role checks a different aspect of the candidate and records its judgment and source-based grounds.

\paragraph{Support: check the observation against media and annotations.}
For event $i$, the Support role receives the source excerpt, cue definition, and cue annotations:
\begin{equation}
R_i^S=\mathcal A_S(X_{q,\tau_i}^{m_i},c_i,\mathbf F_i)
       =(v_i^S,O_i^S,G_i^S).
\label{eq:support}
\end{equation}
$O_i^S$ describes the observed phenomenon. The judgment $v_i^S$ states whether the source supports that phenomenon under the cue definition. $G_i^S$ gives the justification. This keeps the observation separate from the review judgment.

\paragraph{Counter: test local alternatives.}
The Counter role receives the Support record and an expanded interval $\tau_i^+$:
\begin{equation}
R_i^C=\mathcal A_C\big(X_{q,\tau_i^+}^{m_i},c_i,\mathbf F_i,R_i^S,\mathbf G_i^C\big).
\label{eq:counter}
\end{equation}
The context card $\mathbf G_i^C$ compares the core interval with its surroundings. It also supplies within-recording baselines and observation-quality information. Counter checks whether the pattern persists and differs from the person's baseline. It then examines source or annotation inconsistencies as alternative explanations. The output records the counter-evidence relation, issue type, and observation. Technical failure is recorded as unavailable review rather than evidence against the behavior.

\paragraph{Evidence: adjudicate the cue record.}
For cue $c$, the Evidence role considers its candidate records $\mathcal H_c$ and both review sets:
\begin{equation}
 a_c^E=\mathcal A_E(\mathcal H_c,\mathcal R_c^S,\mathcal R_c^C)
       \in\{\mathrm{admit},\mathrm{block}\}.
\label{eq:admission}
\end{equation}
Evidence checks whether support matches the cue definition and source. It also decides whether the challenge weakens that interpretation. Records for the same cue are reviewed together. One problematic interval therefore need not invalidate all other occurrences. The output retains the grounds for admission or exclusion. The supplement shows a schematic example.

The extended Evidence configuration uses all three review roles. Building on this evidence-construction design, the optimized Support policy retains positively verified records without the later challenge and adjudication stages. It preserves the source interval, measurements, Support judgment, and verification grounds. Thus, the final EDA Support model keeps an inspectable evidence record while simplifying admission. We compare the two policies to assess their downstream utility.

\subsection{From Cue Records to Behavioral Assessment}
\label{sec:comparison}

After admission, the behavioral workflow summarizes the retained records for comparison at the recording level. It uses aggregate descriptors and temporal events. For cue $c$ in recording $q$, a sample descriptor $\mathbf f_{q,c}^{S}$ summarizes annotation count, event frequency, duration, interval coverage, and inter-event spacing. The event representation
\begin{equation}
\mathcal G_{q,c}=\{(t_i^s,t_i^e,\mathbf f_i)\}_{i=1}^{N_{q,c}}
\label{eq:eventset}
\end{equation}
retains each event's boundaries and annotation-derived descriptor. Aggregate descriptors summarize overall cue patterns for class-reference comparison. Temporal events retain occurrence times, durations, and links to local context. These fields support source inspection and can contribute complementary comparison information. Their predictive value is evaluated empirically. We compare \emph{Statistical Descriptors}, \emph{Temporal Events}, and \emph{Joint Statistical--Event Representation}. They isolate aggregate descriptors, localized temporal events, and their combined use, respectively.

The reference pool contains same-language recordings from both classes. It excludes the query and its label. Benchmark test queries use training references only. Leave-Query-Out also permits other eligible labeled recordings, as defined in Sec.~\ref{sec:experimental-setup}. Reference membership and normalization change for each query. Model parameters remain frozen.

Statistical distances use robust scaling based on reference medians. Symmetric event-set distances compare timing and annotation-derived features. The supplement specifies the scaling, event comparison, and missing-comparison rules.

For modality setting $M\in\{A,V,AV\}$, let $\mathcal C_M(q)$ contain cues with valid information. Let $\mathcal B_{q,c}$ contain feature blocks that allow comparison with both classes. The class distance is
\begin{equation}
D_y^M(q)=\sqrt{\frac{1}{|\mathcal C_M(q)|}
\sum_{c\in\mathcal C_M(q)}\frac{1}{|\mathcal B_{q,c}|}
\sum_{b\in\mathcal B_{q,c}}\delta_{y,c,b}(q)}.
\label{eq:distance}
\end{equation}
Here $\delta_{y,c,b}$ is the statistical or event-level contribution of block $b$. We average valid blocks within each cue, then average across cues with equal cue weights. The prediction is $\hat y_M=\mathbf1[D_1^M(q)<D_0^M(q)]$. Distance ties go to class $0$. Missing blocks are excluded. If no valid two-class comparison is available, we record an abstention or completion outcome separately.

\subsection{Assessment Matching and Conflict Review}
\label{sec:reconciliation}

EDA compares the text and behavioral predictions. If $\hat y_T=\hat y_{AV}$, it retains their shared label. Otherwise, a text reviewer examines the original transcript, its context, and two anonymized textual judgments. It receives neither the AV label nor the dataset label. It returns a revised label $\hat y_R$, self-reported confidence $c$ in the reviewed label, a rationale, and original-text citations. The behavioral branch determines \emph{when} to review. It does not supply the answer to the reviewer.

For a conflict, the external gate selects the reviewed text judgment when its confidence exceeds 0.8. Otherwise, it uses the AV assessment:
\begin{equation}
\hat y=\begin{cases}
\hat y_R,&v_R=1,\ c>0.8,\\
\hat y_{AV},&\text{otherwise}.
\end{cases}
\label{eq:gate}
\end{equation}
The validity indicator $v_R$ requires an admissible review citing the original text. The reviewed label need not agree with AV. The value $c$ is confidence in that reviewed label, not the probability of depression, and is not assumed to be calibrated. We measure beneficial and harmful revisions and compare them with retaining T. Missing-text completion is analyzed separately to distinguish its effect from that of behavioral evidence.

\section{Experiments}
\label{sec:experiments}

\subsection{Datasets, Reference Construction, and Metrics}
\label{sec:experimental-setup}

\paragraph{Evaluation protocols.}
We evaluate on D-Vlog and LMVD~\cite{yoon2022dvlog,he2026lmvd}. D-Vlog test follows the original benchmark split. Its reference pool uses labeled training recordings only and excludes validation and test recordings. For LMVD, we sort source identifiers and apply stratified ten-fold evaluation with shuffle and seed 42. This follows the documented BHDD-style construction~\cite{ma2026bhdd}. Each fold builds its references and normalization from the other nine folds. We pool held-out predictions to compute the metrics.

We also evaluate D-Vlog under \textbf{Leave-Query-Out}. The reference pool uses other eligible same-language recordings and their labels. It excludes the assessed recording and its label. This is \emph{leave-one-recording-out reference construction}, not leave-one-query-out cross-validation. The model stays frozen and is not retrained for each recording. Only reference membership and derived statistics change. This setting permits labels of other recordings in the assessed cohort. Benchmark testing permits training labels only. We report the two settings separately. \textbf{Consistent with DepressionAgent}~\cite{zhu2026depressionagent}, D-Vlog workflow and ablation tables prioritize Leave-Query-Out. Table~\ref{tab:published} and the supplement retain test comparisons with training references only.

\paragraph{Human cue-annotation check.}
Three affective-computing researchers independently annotate shared 3-second source segments without seeing AI outputs. Unanimous cue labels are retained. Disputed cues are reviewed with two psychological counseling instructors and resolved by five-person binary majority voting. AI cue annotations achieve 91.02\% accuracy against the human labels, pooled across D-Vlog and LMVD. This measures cue agreement on common segments, separately from classification. The supplementary human-validation protocol gives the voting procedure and metric scope.

\paragraph{Backbones and metrics.}
The primary implementation uses frozen Qwen2.5-Omni-7B~\cite{xu2025qwenomni}. Evidence denotes the extended support--challenge--adjudication configuration. The final EDA Support model optimizes admission while retaining source verification and the shared downstream workflow. We compare direct prediction, DepressionAgent~\cite{zhu2026depressionagent}, and EDA at 7B and 30B-A3B. The 7B DepressionAgent row adapts its workflow to the smaller backbone. Metrics are accuracy, precision, recall, and F1 with depression as the positive class. Unclassified outputs count as incorrect in accuracy. All differences are percentage points. Best classification scores are bold within comparable groups, including ties. Dashes denote unreported results.

\begin{table*}[t]

\centering\small\setlength{\tabcolsep}{3pt}\renewcommand{\arraystretch}{1.1}
\caption{\textbf{Matched-test comparison with published methods (\%).}}\label{tab:published}
\begin{tabular*}{\textwidth}{@{\extracolsep{\fill}}llrrrrrrrr@{}}\toprule
 & & \multicolumn{4}{c}{D-Vlog test} & \multicolumn{4}{c}{LMVD}\\
\cmidrule(lr){3-6}\cmidrule(l){7-10}
Method & Year & Acc. & Prec. & Rec. & F1 & Acc. & Prec. & Rec. & F1\\
\midrule
Depression Detector~\cite{yoon2022dvlog} & 2022 & -- & 65.40 & 65.57 & 63.50 & -- & -- & -- & --\\
TAMFN~\cite{zhou2023tamfn} & 2023 & -- & 66.02 & 66.50 & 65.82 & -- & -- & -- & --\\
Attention-MIL~\cite{shangguan2024attentionmil} & 2024 & -- & 67.27 & 67.77 & 66.64 & -- & -- & -- & --\\
Depressformer~\cite{he2024depressformer} & 2024 & 65.00 & 64.00 & 54.00 & 59.00 & -- & -- & -- & --\\
DepMamba~\cite{ye2024depmamba} & 2025 & 68.87 & 68.19 & 86.99 & 76.44 & 72.13 & 70.18 & 76.56 & 73.20\\
CMF-Mamba~\cite{zhou2025cmfmamba} & 2025 & -- & -- & -- & -- & 74.86 & 73.20 & 78.02 & 75.53\\
STCM-Mamba~\cite{zhou2025stcmmamba} & 2025 & 72.64 & 73.72 & 82.11 & 77.69 & 75.41 & 72.55 & 76.68 & 76.49\\
MSI-Net~\cite{shi2026msinet} & 2026 & 70.25 & 78.91 & 69.63 & 73.91 & 74.34 & 78.80 & 75.29 & 77.06\\
MDDFormer~\cite{he2026lmvd} & 2026 & -- & -- & -- & -- & 76.88 & 77.02 & 76.88 & 76.85\\
BHDD~\cite{ma2026bhdd} & 2026 & -- & 75.98 & 75.96 & 75.94 & 78.11 & 78.39 & 78.11 & 78.06\\
DepressionAgent~\cite{zhu2026depressionagent} & 2026 & 95.26 & 95.24 & \textbf{96.15} & \textbf{95.69} & 82.57 & 80.48 & \textbf{85.76} & \textbf{83.04}\\
\midrule
\textbf{EDA Support (7B)} & -- & \textbf{95.32} & \textbf{97.67} & 93.33 & 95.45 & \textbf{83.10} & \textbf{85.38} & 79.74 & 82.46\\

\bottomrule\end{tabular*}\end{table*}

\begin{table}[t]
\centering\small\setlength{\tabcolsep}{2pt}\renewcommand{\arraystretch}{1.0}
\caption{\textbf{Backbone and workflow comparisons (\%).}}\label{tab:matched}
\begin{tabular*}{\linewidth}{@{\extracolsep{\fill}}llrrrr@{}}\toprule
 & & \multicolumn{2}{c}{\shortstack{D-Vlog\\Leave-Query-Out}} & \multicolumn{2}{c}{LMVD}\\
\cmidrule(lr){3-4}\cmidrule(l){5-6}
Backbone & Workflow & Acc. & F1 & Acc. & F1\\
\midrule
7B & Direct & 87.56 & 86.89 & 76.41 & 72.58\\
7B & DepressionAgent & 86.79 & 88.36 & 76.74 & 77.18\\
7B & \textbf{EDA Support} & \textbf{94.30} & \textbf{94.50} & \textbf{83.10} & 82.46\\
7B & EDA Evidence & 94.17 & 94.38 & 82.89 & 82.21\\
30B-A3B & Direct & 81.35 & 84.87 & 74.60 & 79.06\\
30B-A3B & EDA Evidence & 90.16 & 91.26 & 80.53 & 80.40\\
30B-A3B & DepressionAgent & 92.43 & 93.20 & 82.57 & \textbf{83.04}\\
\bottomrule\end{tabular*}\end{table}

\begin{table}[t]
\centering\small\setlength{\tabcolsep}{2pt}\renewcommand{\arraystretch}{1.0}
\caption{\textbf{Statistical and temporal AV representations (\%).}}\label{tab:representation}
\begin{tabular*}{\linewidth}{@{\extracolsep{\fill}}lrrrr@{}}\toprule
 & \multicolumn{2}{c}{\shortstack{D-Vlog\\Leave-Query-Out}} & \multicolumn{2}{c}{LMVD}\\
\cmidrule(lr){2-3}\cmidrule(l){4-5}
Representation & Acc. & F1 & Acc. & F1\\
\midrule
Statistical Descriptors & \textbf{67.36} & \textbf{70.35} & 69.28 & 69.47\\
Temporal Events & 59.97 & 63.86 & 62.26 & 66.01\\
\begin{tabular}[c]{@{}l@{}}Joint Statistical--Event\\Representation\end{tabular} & 67.10 & 70.26 & \textbf{69.56} & \textbf{70.27}\\
\bottomrule\end{tabular*}\end{table}

\begin{table}[t]
\centering\small\setlength{\tabcolsep}{3pt}\renewcommand{\arraystretch}{1.06}
\caption{\textbf{Cue retention and assessment utility (\%).}}\label{tab:ablations}
\begin{tabular*}{\linewidth}{@{\extracolsep{\fill}}lrrr@{}}\toprule
Admission & Cue Retention (\%) & AV F1 & AVT F1\\
\midrule
\multicolumn{4}{@{}l}{\textit{D-Vlog Leave-Query-Out}}\\
Evidence review & 76.38 & 70.26 & 94.38\\
\textbf{EDA Support} & 77.64 & \textbf{71.63} & \textbf{94.50}\\
\midrule
\multicolumn{4}{@{}l}{\textit{LMVD}}\\
Evidence review & 45.94 & 70.27 & 82.21\\
\textbf{EDA Support} & 53.34 & \textbf{70.35} & \textbf{82.46}\\
\bottomrule\end{tabular*}
\end{table}

\subsection{Observe: Behavioral Cue Characterization}
\label{sec:observation-results}
Cue annotations replace a video-level account with named records and source intervals. Figure~\ref{fig:cue-counts} shows that cue occurrence has different associations with depression across datasets. Retained downward head posture (V05) appears in 92.86\% of depression-labeled and 81.97\% of non-depression-labeled D-Vlog recordings. The corresponding LMVD rates are 21.81\% and 21.97\%. Limited head movement (V07) also shows a larger label-group difference on D-Vlog. These differences make cue interpretation dataset-dependent. EDA preserves each cue and its source interval for reference comparison, rather than assigning it a fixed depression meaning. The retained inventory does not itself measure annotation correctness.

\subsection{Verify: Source Grounding and Evidence Admission}
\label{sec:evidence-quality}
\label{sec:ablation}
\textbf{Evidence quality} concerns whether a cue has observable support in its cited source segment. The pooled 91.02\% human-referenced cue accuracy checks this agreement. The shared segmentation supplies the source anchors, so this score does not measure independently estimated boundary accuracy. It also does not establish that one admission policy produces more reliable evidence than another.

\textbf{Assessment utility} concerns what the evidence and its use contribute to classification. Table~\ref{tab:ablations} compares Support with extended Evidence review. Cue Retention is the share of initial candidate cue intervals retained by a policy. It measures admission selectivity, not evidence correctness. Support retains 1.26 points more intervals on D-Vlog Leave-Query-Out and 7.40 points more on LMVD. Its AV F1 gains are 1.37 and 0.08 points, while its AVT F1 gains are 0.12 and 0.25 points. The larger retention change on LMVD accompanies the smaller AV gain. Thus, retention and assessment utility need separate evaluation. Support improves prediction while preserving the shared source-verification mechanism.

\subsection{Assess: Evidence-Guided Depression Assessment}
\label{sec:published-comparison}
\label{sec:matched-results}
\paragraph{Comparisons within the 7B backbone.}
At 7B scale (Table~\ref{tab:matched}), Support improves F1 over direct prediction by 7.61 points on D-Vlog Leave-Query-Out and 9.88 on LMVD. It exceeds 7B DepressionAgent by 6.14 and 5.28 points. Extended Evidence also gains 7.49 and 9.63 points over direct prediction. Thus, the shared evidence-construction framework improves assessment at fixed model scale, with further gains from Support admission.

\paragraph{Reference-setting and benchmark comparisons.}
On D-Vlog Leave-Query-Out, 7B Support reaches 94.30\% accuracy and 94.50\% F1, exceeding 30B-A3B DepressionAgent by 1.87 and 1.30 points. Its AV accuracy is 68.91\%, compared with 63.84\% for DepressionAgent. Table~\ref{tab:published} separately reports matched benchmark comparisons. Support gains 0.06 accuracy points and 2.43 precision points on D-Vlog test, and 0.53 and 4.90 on LMVD. Recall and F1 are lower on both benchmarks. Thus, the precision gains involve a recall trade-off. They should be distinguished from the F1 gains under Leave-Query-Out. The supplement retains complete metrics.

\paragraph{Backbone comparison.}
On D-Vlog Leave-Query-Out, 30B-A3B EDA gains 8.81 accuracy and 6.39 F1 points over direct prediction. However, 7B EDA with the same Evidence policy scores 3.12 F1 points higher. Model scale alone therefore does not determine assessment performance.

\paragraph{Aggregate and temporal evidence.}
\label{sec:modality-results}
Table~\ref{tab:representation} fixes the Evidence-admitted AV events. Statistical Descriptors exceed the joint representation by 0.09 F1 points on D-Vlog Leave-Query-Out, while the joint representation gains 0.80 points on LMVD. Joint representation exceeds Temporal Events on both datasets. Descriptors are therefore strong standalone predictors, while event-level gains depend on the setting. The views also serve different purposes: descriptors summarize overall behavior for class-reference comparison; events preserve occurrence times, duration, and local context for inspection.

\begin{table}[t]
\centering\small\setlength{\tabcolsep}{2pt}\renewcommand{\arraystretch}{1.04}
\caption{\textbf{Review triggering before the final gate (\%).}}\label{tab:review-triggers}
\begin{tabular*}{\linewidth}{@{\extracolsep{\fill}}lrrrr@{}}\toprule
 & \multicolumn{2}{c}{\shortstack{D-Vlog\\Leave-Query-Out}} & \multicolumn{2}{c}{LMVD}\\
\cmidrule(lr){2-3}\cmidrule(l){4-5}
Trigger & Acc. & F1 & Acc. & F1\\
\midrule
Completion & 93.52 & 93.69 & 82.72 & 81.91\\
Random & \shortstack{93.91\\$\pm0.09$} & \shortstack{94.08\\$\pm0.10$} & \shortstack{82.47\\$\pm0.09$} & \shortstack{81.60\\$\pm0.10$}\\
Uncertainty & 93.52 & 93.69 & 82.72 & 81.78\\
AV disagreement & \textbf{94.17} & \textbf{94.38} & \textbf{83.16} & \textbf{82.51}\\
\bottomrule\end{tabular*}
\end{table}

\begin{figure*}[t]
\centering
\includegraphics[width=\textwidth]{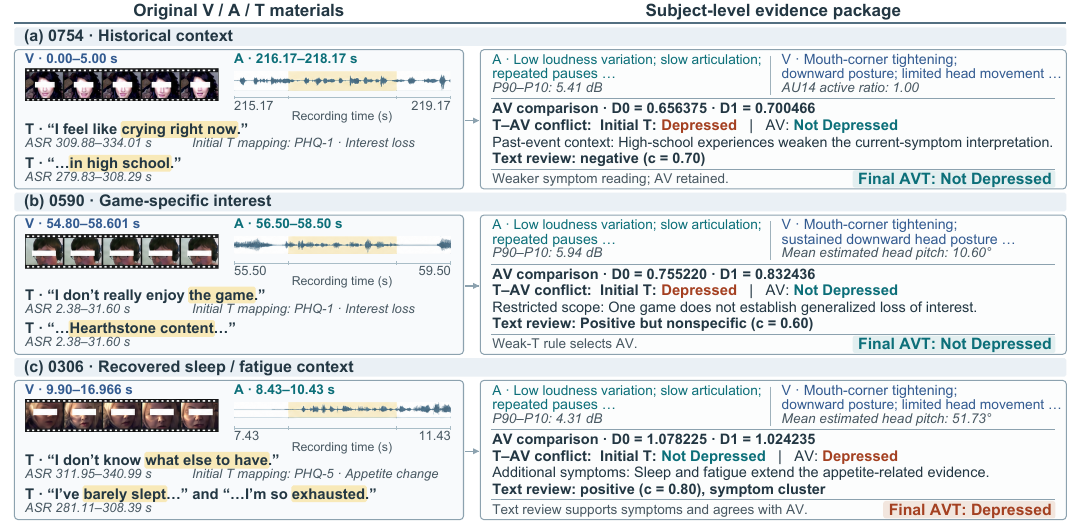}
\caption{\textbf{Source-to-assessment case visualizations.}}
\label{fig:case-traces}
\end{figure*}

\paragraph{Effect of review triggering strategies.}
Table~\ref{tab:review-triggers} compares triggers under matched review budgets, using the same completed text predictions and review rule. Scores precede the final gate; Random reports the mean and standard deviation over five seeds. AV disagreement exceeds Random by 0.26 accuracy and 0.30 F1 points on D-Vlog Leave-Query-Out, and by 0.69 and 0.91 points on LMVD. It also outperforms text-uncertainty triggering. Thus, cross-modal disagreement provides a more informative review-selection signal in these settings. This comparison concerns which predictions to review, separately from how review outcomes enter the final assessment.

\paragraph{Conflict review and accepted changes.}
On LMVD, completion raises accuracy from 81.51\% to 82.72\%. Source review reaches 83.16\%, while the final gate yields 83.10\%. Thus, review selection and final acceptance have distinct effects; the gate does not further improve accuracy here. Among changed predictions, 87.50\% are beneficial on D-Vlog Leave-Query-Out and 58.97\% on LMVD; harmful revisions account for 12.50\% and 41.03\%. These conditional rates differ from overall accuracy gains. The supplement reports uncertainty intervals. Section~\ref{sec:case-studies} traces source context and reported decisions.

\section{Case Studies}
\label{sec:case-studies}
Figure~\ref{fig:case-traces} links source excerpts to subject-level evidence packages for three cases. Each panel shows video frames, an audio excerpt, and transcript passages alongside behavioral measurements, class-reference distances, textual review, and the reported final assessment. The excerpts retain their respective timestamps rather than implying temporal alignment across modalities.

The cases illustrate how textual context affects symptom interpretation. In case (a), reference to high-school experiences weakens the interpretation of the passage as evidence of current symptoms. In case (b), the stated loss of interest concerns a particular game, limiting its support for generalized loss of interest. In case (c), additional sleep and fatigue complaints broaden the evidence beyond the initial appetite-related passage, and the reviewed textual judgment agrees with the behavioral assessment. The displayed review confidences are $c=0.70$, $0.60$, and $0.80$. Each follows the AV fallback under the $c>0.8$ selection rule. These examples make source observations, context, and reported decisions inspectable. They do not establish classification accuracy.

\section{Conclusion}
\label{sec:conclusion}
EDA addresses the difficulty of inspecting the source and use of recording-level behavioral descriptions through three stages. \textbf{Observe} operationalizes five acoustic and four visual cues with literature-informed, label-independent criteria. Evidence Records retain temporal anchors and behavioral measurements. \textbf{Verify} checks observations against their original media and preserves admission grounds. The simplified Support policy retains positively verified records with the same source traceability. \textbf{Assess} combines behavioral reference comparison with PHQ-9-guided text analysis and disagreement-triggered review. On D-Vlog and LMVD, we evaluate cue agreement and downstream utility separately. Cue annotations achieve 91.02\% human-referenced accuracy; at 7B scale, Support improves F1 over direct prediction by 7.61 points under D-Vlog Leave-Query-Out and 9.88 on LMVD. Statistical descriptors are strong predictors, while event-level gains vary by setting. Matched-budget controls support disagreement as a useful review-selection signal. Beyond classification scores, traceable records show how observations enter assessment. Source intervals, measurements, and review grounds let readers distinguish behavioral support from symptom interpretation and inspect unresolved conflicts. These findings encourage explicit analysis and verification of source-media cues alongside representation learning and multimodal fusion. Evidence agreement and prediction performance remain distinct evaluation targets.

\FloatBarrier
\par\label{main:end}
\appendix
\setcounter{section}{0}\setcounter{table}{0}\setcounter{figure}{0}\setcounter{equation}{0}
\renewcommand{\thesection}{S\arabic{section}}
\renewcommand{\thetable}{S\arabic{table}}
\renewcommand{\thefigure}{S\arabic{figure}}
\renewcommand{\theequation}{S\arabic{equation}}
\renewcommand{\theHsection}{supp.\arabic{section}}
\renewcommand{\theHtable}{supp.\arabic{table}}
\renewcommand{\theHfigure}{supp.\arabic{figure}}
\renewcommand{\theHequation}{supp.\arabic{equation}}
\section*{Supplementary Material}
\section{Operational Cue Definitions}
The inventory contains five acoustic and four visual cues from prior behavioral studies. Table~\ref{tab:cue-definitions} lists each cue's observable criterion and literature basis. The definitions describe behavior without using the final depression label. Cue identifiers stay consistent across annotations and experiments.
\begin{table*}[!tbp]

\centering\small\setlength{\tabcolsep}{4pt}\renewcommand{\arraystretch}{1.1}
\caption{\textbf{Literature-informed acoustic and visual cue definitions.}}
\label{tab:cue-definitions}
\begin{tabular*}{\textwidth}{@{\extracolsep{\fill}}lp{.23\textwidth}p{.40\textwidth}p{.23\textwidth}@{}}
\toprule
ID & Observable cue & Annotation criterion & Literature basis\\\midrule
A01 & Low loudness variation & Limited fluctuation in vocal intensity within a localized speech interval & Vocal emphasis/prosody~\cite{gumus2023daily}\\
A02 & Slow pitch change & Gradual pitch changes across voiced speech intervals & Pitch dynamics~\cite{nilsonne1988rate,nilsonne1988speech}\\
A03 & Slow articulation & Reduced articulation pace in an observable speech segment & Speech timing~\cite{cummins2023multilingual}\\
A04 & Long pause & A sustained silence between speech segments, interpreted in speech context & Pause timing~\cite{yamamoto2020timing,mundt2007voice}\\
A05 & Repeated long pauses & Multiple prolonged pauses grouped within a local speech context & Pausing/fluency~\cite{aldeneh2019mood}\\
V01 & Infrequent lip-corner lowering & Low occurrence of visibly lowered lip corners across valid facial observations & Facial activity~\cite{girard2014withdrawal}\\
V03 & Frequent lip-corner tightening & Repeated visible tightening of the lip corners within the local context & Facial activity~\cite{girard2014withdrawal}\\
V05 & Downward head posture & Sustained downward head orientation in an observable interval & Head/gaze orientation~\cite{fiquer2018nonverbal}\\
V07 & Limited head movement & Restricted variation in head orientation across a localized interval & Head-motion dynamics~\cite{girard2014withdrawal}\\
\bottomrule
\end{tabular*}
\end{table*}

\subsection{Annotation Records and Dataset Variation}
Each cue has an observable criterion, a temporal scope, source-validity checks, and interval-merging rules. Each annotation records the recording and cue identifiers, onset, offset, and source excerpt. We derive descriptors from these records for later comparison. Wider context and within-recording baselines support review. The original nonconsecutive visual identifiers are retained for consistency. Text passages are annotated separately by PHQ-9 domain, temporal status, and semantic relation. These fields distinguish current personal symptoms from history, denial, third-party experience, and general discussion. They describe semantic evidence and do not assign questionnaire scores.

Retained downward head posture (V05) appears in 92.86\% of depression-labeled and 81.97\% of non-depression-labeled D-Vlog recordings. The LMVD rates are 21.81\% and 21.97\%. Limited head movement (V07) also shows greater label-group separation on D-Vlog. These patterns describe annotations retained by Evidence. They do not measure exhaustive cue prevalence or clinical specificity. Excluding an annotation does not establish that the cue is absent. Cue-detection recall requires exhaustive human annotations.
\paragraph{Annotation tools and workflow.}
We use automatic feature extraction, candidate interval localization, and multi-agent source-evidence review. Visual features are extracted using OpenFace, including facial action units AU15 and AU14, face-tracking confidence, and head pitch and yaw angles. Audio is decoded into a 16\,kHz mono signal. FSMN-VAD identifies speech activity intervals, from which we extract RMS loudness, fundamental-frequency trajectories, speech duration, and pause information. Table~\ref{tab:cue-measurements} summarizes the annotation and measurement procedures for each cue.

\begin{table*}[!tbp]

\centering\small
\setlength{\tabcolsep}{4pt}
\renewcommand{\arraystretch}{1.1}
\caption{\textbf{Behavioral annotation and measurement procedures.}}
\label{tab:cue-measurements}
\begin{tabular*}{\textwidth}{@{\extracolsep{\fill}}p{.25\textwidth}p{.71\textwidth}@{}}
\toprule
Cue & Behavioral annotation or measurement\\
\midrule
A01: Low loudness variation & Compute RMS loudness in dBFS and summarize its P90--P10 range, standard deviation, interquartile range, and adjacent changes to localize intervals with limited loudness variation.\\
A02: Slow pitch change & Compute the absolute difference between adjacent valid F0 frames, normalized by their mean F0 and temporal separation, to obtain the relative F0 change rate (\%/s) and localize intervals with slow changes.\\
A03: Slow articulation & Divide the number of linguistic units by VAD speech duration to obtain an articulation-rate proxy. English uses words and Chinese uses Han characters; low-rate intervals are localized within the same language and counting unit.\\
A04: Long pause & Use speech activity boundaries to identify silence between adjacent speech segments within the same response, retaining pauses lasting at least 0.5\,s.\\
A05: Repeated long pauses & Reuse A04 pause events to localize at least two long pauses separated by valid speech within the same response.\\
V01: Infrequent lip-corner lowering & Compute the proportion of AU15-positive valid frames among all valid AU15 frames to localize intervals with a low occurrence proportion of lip-corner lowering.\\
V03: High lip-corner tightening proportion & Compute the proportion of AU14-positive valid frames among all valid AU14 frames to localize intervals with a high occurrence proportion of lip-corner tightening.\\
V05: Sustained downward head posture & Use frame-level pitch angles and the within-recording posture baseline to localize candidate downward-head-posture intervals lasting at least 1.5\,s.\\
V07: Limited head movement & Compute angular-amplitude RMS and angular-velocity RMS separately for pitch and yaw to localize intervals with low movement amplitude and speed in both directions, lasting at least 4\,s after merging.\\
\bottomrule
\end{tabular*}
\end{table*}

\section{Evidence Records and Numerical Specification}
\label{sec:supp-numerical}

This section specifies evidence records, cue measurements, source-review context, and reference-comparison rules.

\paragraph{Event provenance.}
Each acoustic or visual candidate stores its recording identifier, modality, cue identifier, start and end times, and source-media locator. It also stores the computed feature vector and available review outputs. Review records distinguish support, counter-evidence, admission, and technical failure. Unavailable review does not count as a negative observation. Source identifiers can link to restricted recordings without making the media public.

\paragraph{Cue measurement and candidate localization.}
We obtain behavioral measurements for the nine cues using the feature extraction procedure described above and localize candidate intervals using the criteria in Table~\ref{tab:cue-selection-parameters}. Candidates then undergo support review, and retained events form the joint statistical--event representation.

\begin{table*}[!tbp]

\centering\small
\setlength{\tabcolsep}{4pt}
\renewcommand{\arraystretch}{1.12}
\caption{\textbf{Measurement criteria and candidate localization parameters for the nine behavioral cues.}}
\label{tab:cue-selection-parameters}
\begin{tabular*}{\textwidth}{@{\extracolsep{\fill}}p{.19\textwidth}p{.47\textwidth}p{.29\textwidth}@{}}
\toprule
Cue & Measurement and candidate threshold & Valid-observation criteria\\
\midrule
A01: Low loudness variation & Loudness P90--P10 $\leq\min(P_{25},8\,\mathrm{dB})$; SD $\leq3$\,dB; IQR $\leq4$\,dB; maximum adjacent change $\leq6$\,dB & Speech proportion $\geq0.7$\\
A02: Slow pitch change & Mean relative F0 change rate $\leq P_{25}$ (\%/s) & Valid-F0 proportion $\geq0.6$; at least 3 baseline windows\\
A03: Slow articulation & Linguistic units per VAD speech second $\leq P_{25}$; candidate duration $\geq1$\,s & Voiced duration $\geq0.8$\,s; at least 3 units; alignment quality $\geq0.7$; at least 3 baseline intervals with the same language and counting unit\\
A04: Long pause & Within-response pause $\geq0.5$\,s & RMS-confirmed silence $\geq0.5$\,s; overlap ratio with VAD silence $\geq0.5$\\
A05: Repeated long pauses & At least 2 A04 pauses within one response; intervening speech $\geq0.3$\,s; local candidate duration $\leq15$\,s & Response duration $\geq2$\,s\\
V01: Infrequent lip-corner lowering & AU15-positive valid-frame proportion $\leq\min(P_{25},0.1)$ & Valid-face proportion $\geq0.8$\\
V03: High lip-corner tightening proportion & AU14-positive valid-frame proportion $\geq\max(P_{60},0.2)$ & Valid-face proportion $\geq0.8$\\
V05: Sustained downward head posture & Pitch above the within-video mean plus one SD; event mean pitch exceeds this threshold by at least $1^\circ$; duration $\geq1.5$\,s & Tracking confidence $\geq0.8$; adjacent observation gap $\leq0.4$\,s\\
V07: Limited head movement & Pitch and yaw amplitude RMS each $\leq1^\circ$; angular-velocity RMS each $\leq15^\circ$/s; joint RMS of the four median-normalized measurements $\leq P_{25}$; merged duration $\geq4$\,s & Valid-pose proportion $\geq0.8$; at least 3 frames\\
\bottomrule
\end{tabular*}
\par\smallskip
\begin{minipage}{\textwidth}\footnotesize
\textit{Note:} $P_{25}$ and $P_{60}$ are the 25th and 60th percentiles of each cue's measurements over valid observations within the video. Visual tracking confidence must be at least 0.8.
\end{minipage}
\end{table*}

\paragraph{Joint Statistical--Event Representation.}
We retain sample-level statistics and event-level measurements as complementary views of the same evidence. Statistics summarize overall behavioral patterns for class-reference comparison. Events preserve local timing, duration, source links, and features for inspection and possible complementary comparison. Table~\ref{tab:joint-cue-measurements} lists the cue-specific measures.

\begin{table*}[!tbp]

\centering\small
\setlength{\tabcolsep}{4pt}
\renewcommand{\arraystretch}{1.12}
\caption{\textbf{Sample-level statistics and event-level measurements for the nine cues.}}
\label{tab:joint-cue-measurements}
\begin{tabular*}{\textwidth}{@{\extracolsep{\fill}}p{.06\textwidth}p{.46\textwidth}p{.43\textwidth}@{}}
\toprule
Cue & Sample-level statistics & Event-level measurements\\
\midrule
A01 & Median of within-window loudness P90--P10 & Loudness P90--P10 (dB)\\
A02 & Median of within-window mean relative F0 change rate & Mean relative F0 change rate (\%/s)\\
A03 & Median articulation rate within the same language and counting unit & Articulation rate (words/s or Han characters/s)\\
A04 & Pause-union duration divided by response-proxy duration; median pause duration; events per minute & Duration only\\
A05 & Pause-union duration divided by response-proxy duration & Pause count; total pause duration divided by event duration\\
V01 & AU15-positive valid-frame proportion & AU15-positive valid-frame proportion\\
V03 & AU14-positive valid-frame proportion & AU14-positive valid-frame proportion\\
V05 & Median duration of downward-head-posture events & Mean downward pitch angle (degrees)\\
V07 & Window medians of pitch amplitude RMS, pitch angular-velocity RMS, yaw amplitude RMS, and yaw angular-velocity RMS & Pitch amplitude RMS, pitch angular-velocity RMS, yaw amplitude RMS, and yaw angular-velocity RMS\\
\bottomrule
\end{tabular*}
\end{table*}

\paragraph{Review context.}
Counter-evidence review uses up to 3\,s of source audio/video before and after each candidate event, clipped to the media boundaries. Event duration and behavioral measurements are computed over the original candidate interval.

\paragraph{Distance aggregation and missing comparisons.}
The event-level discrepancy between a query and a class is the median symmetric distance to reference samples of that class. Measurements are averaged equally within each statistical block. Available statistical and event blocks are then averaged equally within each cue, followed by an equal-weight average across valid cues and a square root to obtain the class distance. Reference comparisons are grouped by language; articulation rate additionally requires the same counting unit. Comparisons involving empty event sets or missing references for either class are excluded from aggregation. Samples with no valid comparison retain an unavailability flag and receive the default class 0 in full-denominator evaluation. Ties between class distances are assigned to class 0.

\paragraph{Model and generation settings.}
Support review uses the frozen Qwen2.5-Omni-7B model with a generation limit of 256 tokens. Independent and direct source-evidence reviews in the held-out experiments use the same model, with initial generation budgets of 512 and 384 tokens and maximum budgets of 2,048 and 768 tokens, respectively.

\subsection{Reference Scaling and Event Comparison}
For numerical coordinate $j$, let $\mu_{yj}$ be the median among references of class $y$. A robust scale fitted over the permitted references is
\begin{equation}
s_j=\max\{Q_{0.75,j}-Q_{0.25,j},\ 0.05|Q_{0.5,j}|,\ 10^{-6}\}.
\end{equation}
The coordinate discrepancy is
\begin{equation}
a_{yj}(q)=\min\!\left(\frac{|f_{qj}-\mu_{yj}|}{s_j},4\right)^2.
\end{equation}
This measures deviation from the class reference after scaling and limits the effect of extreme values. For events, a symmetric discrepancy compares timing and annotation-derived descriptors with each class's references. The median discrepancy across references supplies the event-level contribution.

\section{Textual Evidence and Review Boundaries}
\label{sec:supp-text}

\begin{figure*}[!tp]

\centering
\includegraphics[width=\textwidth]{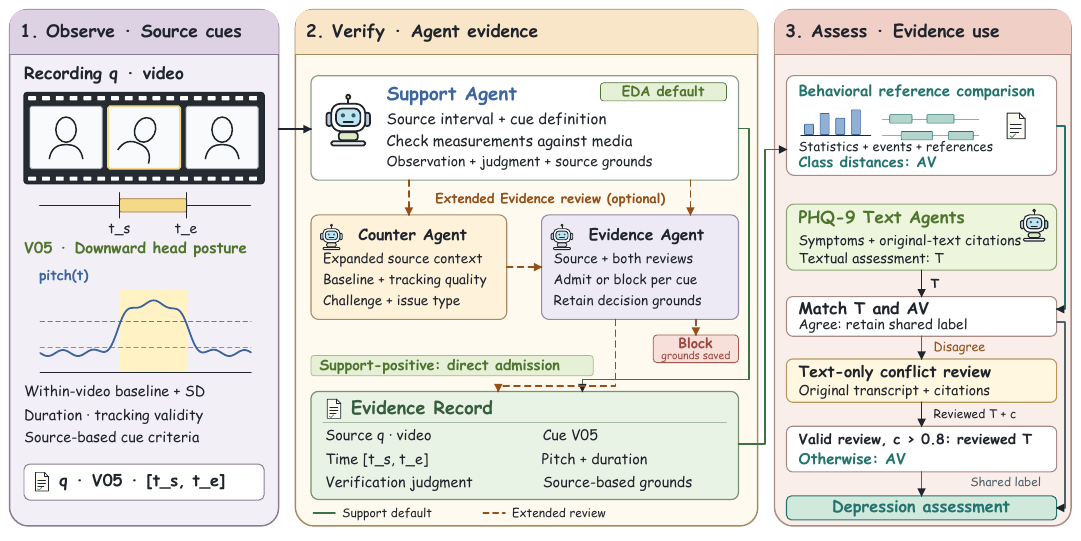}
\caption{\textbf{Source-grounded verification and evidence use.}}
\label{fig:supp-audit}
\end{figure*}

Figure~\ref{fig:supp-audit} traces an illustrative V05 record from recording $q$ and video interval $[t_s,t_e]$ to assessment. Pitch, duration, and tracking validity support source verification; the traces are symbolic, not measured participant data. The solid Support path admits positively verified records directly. Dashed paths add Counter challenge and Evidence adjudication. Retained source anchors, measurements, judgments, and grounds feed behavioral reference comparison. The resulting AV assessment is matched with the independently obtained text assessment; disagreement triggers text-only review and the confidence gate. No single cue directly determines the depression label. Table~\ref{tab:ablation} reports text-analysis stages, and Table~\ref{tab:decision-controls} separates completion, source review, and final acceptance.

\paragraph{Textual evidence organization.}
We organize original transcript passages into PHQ-9 symptom support, explicit denial, personal-status statements, and contextual qualifiers. Each record retains its source identifier and quotation, distinguishing current or temporally unspecified personal experiences from historical states, third-party accounts, and general discussion. PHQ-9 provides symptom categories~\cite{kroenke2001phq}, not questionnaire scores inferred from text. An unmentioned symptom is not treated as a denial.

\paragraph{Textual assessment and internal review.}
Unimodal text analysis comprises evidence extraction, initial assessment, and internal review. Extraction organizes traceable source evidence; initial assessment integrates symptom support, denial, and alternative explanations; internal review independently revisits the source to examine the experiencer, temporal context, and symptom meaning. A valid review changes the initial prediction only when it disagrees with that prediction and its confidence in the selected class is at least 0.8, yielding the final text label T.

\paragraph{Conflict-triggered review.}
When valid T and AV predictions disagree, the conflict-review agent first forms an independent judgment from the source text and then reviews anonymized versions of the original text judgment and the independent judgment.

\paragraph{Output validation and acceptance.}
Conflict review returns a label, rationale, source identifiers, and confidence $c$ in its selected label. If the initial T and AV labels agree, their shared label is retained. Otherwise, a valid reviewed text label is used when $c>0.8$; if $c\leq0.8$ or the review is invalid, AV is used. Validity requires an admissible output with nonempty citations belonging to the current source. A confident reviewed text label can therefore override AV. The reviewer does not receive the AV or ground-truth label. Confidence is self-reported and is not assumed to be calibrated. Missing text falls back to the available AV prediction, and unavailable T and AV predictions fall back to class 0.

Sections~\ref{sec:text-agent-prompts} and~\ref{sec:conflict-agent-prompt} present the prompts and output formats for four roles: textual evidence extraction, initial textual assessment, internal review, and conflict review. Each template specifies the role, inputs, task requirements, decision rules, and expected output.

\section{Complete Reference-setting and Modality Results}
\begin{table*}[!tbp]

\centering\footnotesize\setlength{\tabcolsep}{3pt}\renewcommand{\arraystretch}{1.0}
\caption{\textbf{D-Vlog Leave-Query-Out comparison (\%).}}\label{tab:lqo-detail}
\begin{tabular*}{\textwidth}{@{\extracolsep{\fill}}lrrrr@{}}\toprule
Method & Acc. & Prec. & Rec. & F1\\
\midrule
DepressionAgent (30B-A3B) & 92.43 & 92.73 & \textbf{93.67} & 93.20\\
EDA Support (7B) & \textbf{94.30} & \textbf{95.94} & 93.10 & \textbf{94.50}\\
EDA Evidence (7B) & 94.17 & 95.70 & 93.10 & 94.38\\
\bottomrule\end{tabular*}\end{table*}

\begin{table*}[!tbp]

\centering\footnotesize\setlength{\tabcolsep}{3pt}\renewcommand{\arraystretch}{1.0}
\caption{\textbf{Complete modality results (\%).}}\label{tab:full-scopes}
\begin{tabular*}{\textwidth}{@{\extracolsep{\fill}}llrrrrrr@{}}\toprule
 &  & \multicolumn{2}{c}{\shortstack{D-Vlog\\Leave-Query-Out}} & \multicolumn{2}{c}{D-Vlog test} & \multicolumn{2}{c}{LMVD}\\
System & Modality & Acc. & F1 & Acc. & F1 & Acc. & F1\\
\midrule
DepressionAgent (30B-A3B) & A & 57.13 & 39.75 & -- & -- & -- & --\\
 & V & 44.68 & 0.00 & -- & -- & -- & --\\
 & AV & 63.84 & 54.80 & -- & -- & -- & --\\
 & T & 89.60 & 90.38 & -- & -- & -- & --\\
 & AVT & 92.43 & 93.20 & -- & -- & -- & --\\
EDA Support (7B) & A & \textbf{67.10} & \textbf{68.64} & 65.50 & 68.78 & 69.01 & 69.18\\
 & V & \textbf{61.79} & 64.76 & 63.16 & 64.80 & \textbf{61.22} & \textbf{57.64}\\
 & AV & \textbf{68.91} & \textbf{71.63} & \textbf{71.35} & 74.61 & \textbf{69.67} & \textbf{70.35}\\
 & T & \textbf{93.52} & \textbf{93.69} & \textbf{94.15} & \textbf{94.25} & \textbf{81.51} & \textbf{81.87}\\
 & AVT & \textbf{94.30} & \textbf{94.50} & \textbf{95.32} & \textbf{95.45} & \textbf{83.10} & \textbf{82.46}\\
EDA Evidence (7B) & A & \textbf{67.10} & 68.56 & \textbf{66.08} & \textbf{69.47} & \textbf{69.45} & \textbf{69.58}\\
 & V & 61.53 & \textbf{64.77} & \textbf{64.33} & \textbf{67.03} & 56.83 & 46.50\\
 & AV & 67.10 & 70.26 & \textbf{71.35} & \textbf{74.87} & 69.56 & 70.27\\
 & T & \textbf{93.52} & \textbf{93.69} & \textbf{94.15} & \textbf{94.25} & \textbf{81.51} & \textbf{81.87}\\
 & AVT & 94.17 & 94.38 & \textbf{95.32} & \textbf{95.45} & 82.89 & 82.21\\
\bottomrule\end{tabular*}\end{table*}

Table~\ref{tab:lqo-detail} reports the Leave-Query-Out comparison and adds precision and recall to the main-paper accuracy and F1 results. DepressionAgent values come from its published reference-setting analysis. EDA uses the leave-one-recording-out reference construction defined in the main paper.

Table~\ref{tab:full-scopes} reports A, V, AV, T, and AVT for both admission settings. Support and Evidence use the same text predictions. The main-paper representation controls use the same Evidence-admitted AV events. Only the statistical or temporal representation changes.

\begin{table*}[!tbp]

\centering\footnotesize\setlength{\tabcolsep}{3pt}\renewcommand{\arraystretch}{1.0}
\caption{\textbf{Complete backbone and workflow comparisons (\%).}}\label{tab:backbones-complete}
\begin{tabular*}{\textwidth}{@{\extracolsep{\fill}}llrrrrrr@{}}\toprule
 &  & \multicolumn{2}{c}{\shortstack{D-Vlog\\Leave-Query-Out}} & \multicolumn{2}{c}{D-Vlog test} & \multicolumn{2}{c}{LMVD}\\
\cmidrule(lr){3-4}\cmidrule(lr){5-6}\cmidrule(l){7-8}
Backbone & Workflow & Acc. & F1 & Acc. & F1 & Acc. & F1\\
\midrule
7B & Direct & 87.56 & 86.89 & 88.89 & 88.48 & 76.41 & 72.58\\
7B & DepressionAgent & 86.79 & 88.36 & 87.72 & 89.23 & 76.74 & 77.18\\
7B & \textbf{EDA Support} & \textbf{94.30} & \textbf{94.50} & \textbf{95.32} & 95.45 & \textbf{83.10} & 82.46\\
7B & EDA Evidence & 94.17 & 94.38 & \textbf{95.32} & 95.45 & 82.89 & 82.21\\
30B-A3B & Direct & 81.35 & 84.87 & 83.63 & 86.54 & 74.60 & 79.06\\
30B-A3B & EDA Evidence & 90.16 & 91.26 & 92.40 & 93.19 & 80.53 & 80.40\\
30B-A3B & DepressionAgent & 92.43 & 93.20 & 95.26 & \textbf{95.69} & 82.57 & \textbf{83.04}\\
\bottomrule\end{tabular*}\end{table*}

\begin{table*}[!tbp]

\centering\footnotesize\setlength{\tabcolsep}{3pt}\renewcommand{\arraystretch}{1.0}
\caption{\textbf{Complete AV representation controls (\%).}}\label{tab:representation-complete}
\begin{tabular*}{\textwidth}{@{\extracolsep{\fill}}lrrrrrr@{}}\toprule
 & \multicolumn{2}{c}{\shortstack{D-Vlog\\Leave-Query-Out}} & \multicolumn{2}{c}{D-Vlog test} & \multicolumn{2}{c}{LMVD}\\
\cmidrule(lr){2-3}\cmidrule(lr){4-5}\cmidrule(l){6-7}
Representation & Acc. & F1 & Acc. & F1 & Acc. & F1\\
\midrule
Statistical Descriptors & \textbf{67.36} & \textbf{70.35} & 70.76 & 74.23 & 69.28 & 69.47\\
Temporal Events & 59.97 & 63.86 & 66.08 & 69.79 & 62.26 & 66.01\\
\begin{tabular}[c]{@{}l@{}}Joint Statistical--Event\\Representation\end{tabular} & 67.10 & 70.26 & \textbf{71.35} & \textbf{74.87} & \textbf{69.56} & \textbf{70.27}\\
\bottomrule\end{tabular*}\end{table*}

Table~\ref{tab:backbones-complete} compares backbones and workflows, while Table~\ref{tab:representation-complete} compares statistical, temporal, and joint AV representations.

The main paper prioritizes D-Vlog Leave-Query-Out for workflow, representation, and admission comparisons. These complete tables also report D-Vlog test results with training references only. In Table~\ref{tab:representation-complete}, the joint representation on that test split uses fixed Evidence-admitted inputs. It reaches 71.35\% accuracy and 74.87\% F1. The gains over descriptors alone are 0.59 and 0.64 points. The gains over events alone are 5.27 and 5.08 points. In contrast, descriptors alone score slightly higher under Leave-Query-Out. Combining representations therefore does not help equally across evaluation settings.

Bold marks the best scores, including ties. Modality tables compare systems within each modality. Text-stage tables compare stages within the same dataset and protocol.

\section{Text Analysis and Reference Protocols}
\label{sec:supp-experimental-setup}
\begin{table*}[!tbp]

\centering\small\setlength{\tabcolsep}{3pt}\renewcommand{\arraystretch}{1.1}
\caption{\textbf{Text-analysis stages and final Support assessment (\%).}}\label{tab:ablation}
\begin{tabular*}{\textwidth}{@{\extracolsep{\fill}}lrrrrrr@{}}\toprule
Dataset / protocol & \multicolumn{2}{c}{Initial T} & \multicolumn{2}{c}{Internal T review} & \multicolumn{2}{c}{Final AVT (Support)}\\
 & Acc. & F1 & Acc. & F1 & Acc. & F1\\
\midrule
D-Vlog Leave-Query-Out & 90.28 & 90.54 & 93.52 & 93.69 & \textbf{94.30} & \textbf{94.50}\\
D-Vlog test & 90.64 & 91.01 & 94.15 & 94.25 & \textbf{95.32} & \textbf{95.45}\\
LMVD & 78.55 & 78.28 & 81.51 & 81.87 & \textbf{83.10} & \textbf{82.46}\\
\bottomrule\end{tabular*}\end{table*}

Table~\ref{tab:ablation} reports Support text-analysis stages alongside the modality results. For LMVD, we sort identifiers before stratified ten-fold splitting with shuffle and seed 42. Leave-Query-Out uses other eligible recordings and their labels as references. It excludes the query and its label. The foundation model stays frozen. This is leave-one-recording-out reference construction. D-Vlog test uses original training references only. LMVD builds references and normalization within the training folds, then pools held-out predictions across folds.

Table~\ref{tab:ablation} separates initial text analysis, internal review, and final multimodal assessment; Table~\ref{tab:full-scopes} provides the Support--Evidence modality comparison. On LMVD, Support reaches 83.10\% accuracy and 82.46\% F1. Evidence reaches 82.89\% and 82.21\%. Initial and internally reviewed text predictions stay fixed across admission settings. The final AVT row uses Support admission. Missing-text completion is tested separately in Table~\ref{tab:decision-controls}. The accompanying structured data retain run identifiers, source references, and transcribed values.

\begin{table*}[!tbp]

\centering\small
\renewcommand{\arraystretch}{1.12}
\setlength{\tabcolsep}{3pt}
\caption{\textbf{Behavioral admission on D-Vlog Leave-Query-Out (\%).}}\label{tab:support}
\begin{tabular*}{\textwidth}{@{\extracolsep{\fill}}lr*{8}{r}@{}}\toprule
 & & \multicolumn{2}{c}{A} & \multicolumn{2}{c}{V} & \multicolumn{2}{c}{AV} & \multicolumn{2}{c}{AVT}\\
\cmidrule(lr){3-4}\cmidrule(lr){5-6}\cmidrule(lr){7-8}\cmidrule(l){9-10}
Admission & Intervals & Acc. & F1 & Acc. & F1 & Acc. & F1 & Acc. & F1\\\midrule
Support & 99,166 & \textbf{67.10}& \textbf{68.64}& \textbf{61.79}& 64.76 & \textbf{68.91}& \textbf{71.63}& \textbf{94.30}& \textbf{94.50}\\
+ Evidence & 97,553 & \textbf{67.10}& 68.56 & 61.53 & \textbf{64.77}& 67.10 & 70.26 & 94.17 & 94.38\\
\bottomrule\end{tabular*}

\end{table*}
\begin{table*}[!tbp]

\centering\small\setlength{\tabcolsep}{4pt}
\caption{\textbf{Cue-admission controls on D-Vlog test (\%).}}\label{tab:heldout-admission}
\begin{tabular*}{\textwidth}{@{\extracolsep{\fill}}lrrrrrrrr@{}}\toprule
 & \multicolumn{2}{c}{A} & \multicolumn{2}{c}{V} & \multicolumn{2}{c}{AV} & \multicolumn{2}{c}{AVT}\\
\cmidrule(lr){2-3}\cmidrule(lr){4-5}\cmidrule(lr){6-7}\cmidrule(l){8-9}
Admission & Acc. & F1 & Acc. & F1 & Acc. & F1 & Acc. & F1\\\midrule
Support & 65.50 & 68.78 & 63.16 & 64.80 & \textbf{71.35}& 74.61 & \textbf{95.32}& \textbf{95.45}\\
Evidence & \textbf{66.08}& \textbf{69.47}& \textbf{64.33}& \textbf{67.03}& \textbf{71.35}& \textbf{74.87}& \textbf{95.32}& \textbf{95.45}\\
\bottomrule\end{tabular*}

\end{table*}
\begin{table*}[!tbp]
\centering\small\setlength{\tabcolsep}{5pt}\renewcommand{\arraystretch}{1.08}
\caption{\textbf{Decision controls for the final Support model (\%).}}
\label{tab:reference-decision-controls}
\begin{tabular*}{\textwidth}{@{\extracolsep{\fill}}lrrrr@{}}\toprule
 & \multicolumn{2}{c}{D-Vlog Leave-Query-Out} & \multicolumn{2}{c}{LMVD}\\
\cmidrule(lr){2-3}\cmidrule(l){4-5}
Configuration & Acc. & F1 & Acc. & F1\\\midrule
EDA Support & \textbf{94.30} & \textbf{94.50} & 83.10 & 82.46\\
Without conflict review & 93.52 & 93.69 & 82.72 & 81.91\\
Without confidence gate & 94.17 & 94.38 & \textbf{83.16} & \textbf{82.51}\\
\bottomrule\end{tabular*}
\end{table*}

\begin{table*}[!tbp]

\centering\small\setlength{\tabcolsep}{3pt}\renewcommand{\arraystretch}{1.1}
\caption{\textbf{Completion, review, and acceptance controls (\%).}}\label{tab:decision-controls}
\begin{tabular*}{\textwidth}{@{\extracolsep{\fill}}lrrrrrr@{}}\toprule
Rule & \multicolumn{2}{c}{\shortstack{D-Vlog\\Leave-Query-Out}} & \multicolumn{2}{c}{D-Vlog test} & \multicolumn{2}{c}{LMVD}\\
 & Acc. & F1 & Acc. & F1 & Acc. & F1\\
\midrule
T (unclassified retained) & 93.52 & 93.69 & 94.15 & 94.25 & 81.51 & 81.87\\
+ completion & 93.52 & 93.69 & 94.15 & 94.25 & 82.72 & 81.91\\
+ source review & 94.17 & 94.38 & \textbf{95.32} & \textbf{95.45} & \textbf{83.16} & \textbf{82.51}\\
+ final gate (AVT) & \textbf{94.30} & \textbf{94.50} & \textbf{95.32} & \textbf{95.45} & 83.10 & 82.46\\

\bottomrule\end{tabular*}

\end{table*}
\begin{table*}[!tbp]

\centering\small
\setlength{\tabcolsep}{3pt}
\caption{\textbf{Accepted prediction revisions with Support admission.}}
\label{tab:support-revision-summary}
\begin{tabular*}{\textwidth}{@{\extracolsep{\fill}}lrrrrrrr@{}}
\toprule
Protocol & Samples & Revised & Corrected & Harmed & Revised (\%) & Beneficial (\%, 95\% CI) & $\Delta$Acc. (pp)\\
\midrule
\shortstack[l]{D-Vlog\\Leave-Query-Out} & 772 & 8 & 7 & 1 & 1.04 & 87.50 [52.91, 97.76] & +0.78\\
D-Vlog test & 171 & 2 & 2 & 0 & 1.17 & 100.00 [34.24, 100.00] & +1.17\\
LMVD & 1,823 & 39 & 23 & 16 & 2.14 & 58.97 [43.42, 72.92] & +0.38\\
\bottomrule
\end{tabular*}

\end{table*}

\section{Admission and Decision Details}
\label{sec:supp-controls}

\paragraph{Admission configurations.}
Tables~\ref{tab:support} and~\ref{tab:heldout-admission} compare admission configurations on D-Vlog Leave-Query-Out and D-Vlog test, respectively; Table~\ref{tab:full-scopes} provides the Support--Evidence comparison across modalities and datasets. Support retains positively supported candidates. Evidence weighs support against counter-evidence before admission. It can retain admissible observations despite a challenge. Technical failure is recorded separately from counter-evidence.

On D-Vlog Leave-Query-Out, Support reaches 94.30\% AVT accuracy and Evidence reaches 94.17\%. Support also exceeds Evidence on LMVD. On D-Vlog test, Support and Evidence have the same final AVT scores. Their A, V, and AV results still differ. On LMVD, Support reaches 83.10\% accuracy and Evidence reaches 82.89\%. Thus, changes in behavioral evidence need not produce the same changes in final assessment.

\paragraph{Completion, source review, and acceptance.}

Table~\ref{tab:reference-decision-controls} retains the main model's conflict-review and confidence-gate controls. Table~\ref{tab:decision-controls} further separates completion, source review, and final selection. The source-review and gate rows use the same completion outputs. The gate also reuses the source-review results. On LMVD, completion raises Support accuracy from 81.51\% to 82.72\%. Source review reaches 83.16\%, and the gate yields 83.10\%. The gate improves D-Vlog Leave-Query-Out accuracy by 0.13 points but lowers LMVD accuracy by 0.06 points. Its benefit therefore depends on the dataset. These controls fix Support admission and vary only review and final selection. Table~\ref{tab:support-revision-summary} reports accepted revisions, corrected and harmed predictions, beneficial-revision rates, and overall accuracy changes. Revision-quality rates are conditional on changed outputs. Corrected minus harmed predictions equals the change in correct predictions from completion to final AVT.

\section{Annotation Inventory and Source Duration}
\begin{table*}[!tbp]

\centering\small\setlength{\tabcolsep}{3pt}\renewcommand{\arraystretch}{1.1}
\caption{\textbf{Source-media durations and cue-admission totals.}}\label{tab:annotation-totals}
\begin{tabular*}{\textwidth}{@{\extracolsep{\fill}}lrrrrr@{}}\toprule
Dataset / protocol & Audio hours & Video hours & Candidates & Support & Evidence\\
\midrule
D-Vlog Leave-Query-Out & 123.85 & 123.85 & 127,719 & 99,166 & 97,553\\
D-Vlog test & 25.96 & 25.96 & 27,079 & 20,992 & 20,699\\
LMVD & 236.73 & 214.62 & 193,710 & 103,332 & 88,994\\
\bottomrule\end{tabular*}\end{table*}

Table~\ref{tab:annotation-totals} reports source-media durations and cue-admission totals. Annotation totals count cue-specific intervals. The main-paper distribution figure instead counts each recording once per retained cue. These units are not interchangeable. Source durations count each media input once per modality. Candidates, Support, and Evidence columns report cue-specific interval totals, not recording counts. Cue retention is the retained interval count divided by the initial candidate interval count, multiplied by 100\%. It measures admission selectivity, not evidence correctness. LMVD uses separate audio and video sources, so their total durations differ.

\subsection{Human Annotation and Source-Grounding Validation Protocol}
\label{sec:human-grounding-protocol}
The human evaluation covers annotated material from both D-Vlog and LMVD. Humans and AI use the same pre-segmented source units: consecutive 3-second windows for audio and for video. Annotators select the applicable cues from the literature-informed nine-cue codebook, using the corresponding modality and observable criteria. Adjacent windows with consistent cue assignments are merged into annotation intervals. The common segmentation avoids matching independently chosen human and AI boundaries.

Three researchers in affective computing independently review the source segments without seeing AI annotations. For each segment, each cue receives a binary present/absent judgment. A cue with unanimous initial judgments is retained without further review. Only cues with disagreement are referred to a panel comprising the original three researchers and two instructors in psychological counseling. The instructors participate in reviewing the disputed cues rather than creating a separate annotation set. All five panel members then vote on each disputed cue, and a majority of at least three determines its binary label. Agreement and escalation are therefore assessed per cue, not by comparing entire multi-label sets. The resulting human consensus is the ground-truth basis for evaluating the AI annotations.

We calculate annotation accuracy as
\begin{equation}
\mathrm{Acc}_{\mathrm{cue}}=
\frac{K_{\mathrm{match}}}{K_{\mathrm{eval}}}\times100\%.
\end{equation}
Here $K_{\mathrm{eval}}$ counts all evaluated AI annotations, and $K_{\mathrm{match}}$ counts those matching the human labels. The pooled accuracy is 91.02\% across the two datasets. It is not a D-Vlog-only result, an inter-rater reliability statistic, or a policy-specific Support--Evidence comparison. No separate per-dataset score is inferred from it.

This protocol addresses the three research questions at distinct levels. For \emph{Observe}, the shared codebook makes the nine cue choices explicit. For \emph{Verify}, blinded human judgments test whether the AI cue assignments agree with observations of the same source segments. Their timestamps identify the inspected locations, but the score does not evaluate free-boundary temporal localization or the faithfulness of every generated rationale. For \emph{Assess}, classification and accepted-revision metrics remain separate from annotation accuracy. They measure downstream utility, not the correctness of an individual cue.

\paragraph{Revision interpretation.}
The revision counts and beneficial-revision rates in Table~\ref{tab:support-revision-summary} apply only to outputs that changed. They do not measure correction over the whole cohort. The main discussion therefore reports these rates separately from overall accuracy gains.

\section{Case Studies and Source-to-Assessment Traces}
\label{sec:supp-case-protocol}

Main-paper Figure~4 presents cases 0754, 0590, and 0306. Each case pairs video frames, an audio trace, and transcript passages with a subject-level evidence package. The package shows acoustic and visual measurements, class-reference distances, textual context review, and the reported final assessment. Table~\ref{tab:case-catalog} summarizes the displayed cases.

\par\medskip\noindent\begin{minipage}{\linewidth}
\centering\small\setlength{\tabcolsep}{4pt}
\captionof{table}{\textbf{Cases displayed in the main-paper source trace.}}
\label{tab:case-catalog}
\begin{tabular*}{\columnwidth}{@{\extracolsep{\fill}}p{0.26\columnwidth}p{0.67\columnwidth}@{}}
\toprule
Case & Source context and reported outcome\\\midrule
0754 & High-school context weakens a current-symptom reading; final non-depression assessment.\\
0590 & Game-specific interest limits symptom scope; final non-depression assessment.\\
0306 & Sleep and fatigue add symptom context; review agrees with AV; final depression assessment.\\
\bottomrule
\end{tabular*}
\end{minipage}\par\medskip

\subsection{Source Excerpts and Context}
The video, audio, and text excerpts retain their respective source timestamps. They are selected excerpts from the same recording, and their horizontal placement does not imply temporal alignment. The highlighted passages show why the scope of a textual statement matters. In 0754, high-school experiences weaken a current-symptom interpretation. In 0590, loss of interest concerns one game. In 0306, sleep and fatigue complaints add to the initial appetite-related passage.

\subsection{Behavioral Comparison and Reported Decisions}
The evidence packages display the actual supplied measurements and distances to both class references. The displayed textual review explains how source context changes the interpretation of the initial symptom passage. Cases 0754 and 0590 report a final non-depression assessment, while 0306 reports depression. Their post-review text confidences are $c=0.70$, $0.60$, and $0.80$, respectively. The confidence selector therefore uses AV in all three displayed cases, including the boundary value $c=0.80$. The high-confidence reviewed-text branch is defined in the method. These are the outcomes in the supplied case records. The figure does not assign a ground-truth correctness label or establish the causal contribution of an individual cue. Figure~\ref{fig:supp-audit} separately explains the cue-level verification mechanism as a schematic.

\subsection{Prediction Revision Results}
Table~\ref{tab:support-revision-summary} reports accepted revisions under Support admission. Among changed predictions, 87.50\% are beneficial on D-Vlog Leave-Query-Out and 58.97\% on LMVD. The count-based net revision gains are 0.78 and 0.38 percentage points. D-Vlog test has a beneficial-revision rate of 100.00\% and a net gain of 1.17 points, with the uncertainty interval reported in the table. Revision rates use changed predictions as the denominator, while net accuracy gains use the full evaluated cohort. The counts compare completed text predictions with final AVT predictions, matching the corresponding rows in Table~\ref{tab:decision-controls}.

\section{Agent Prompts and Output Templates}
\label{sec:supp-prompts}
This section presents the text-agent prompts in English, organized by role, inputs, task rules, and output format. The templates translate and organize the source instructions; the evidence-extraction template consolidates its constituent tasks.

\subsection{Cue Annotation}
Candidate intervals are generated by rules and subsequently submitted for evidence review.

\noindent\textbf{Task.} Locate observable acoustic and visual behaviors and record their temporal intervals, measurements, and source references, without predicting depression labels.

\noindent\textbf{Inputs.} Time-stamped acoustic and visual features, media timing information, and the definitions and selection parameters of the corresponding cues. Depression labels are excluded from the inputs.

\noindent\textbf{Annotation rules.}
\begin{itemize}
\setlength{\itemsep}{0pt}\setlength{\parsep}{0pt}
\item Acoustic cues comprise loudness variation (A01), fundamental-frequency variation (A02), articulation rate (A03), pause duration (A04), and repeated pauses (A05). Visual cues comprise AU15 (V01), AU14 (V03), head lowering (V05), and head movement (V07).
\item Apply the measurement definitions, thresholds, and valid-observation rules specified earlier for each cue. Record the corresponding behavioral measurements; overall impressions of affect do not substitute for specific behaviors.
\item Express interval boundaries in seconds on the original media timeline. Merge intervals according to the cue-specific rules and retain source references for their constituent windows.
\item Retain quality flags and measurement evidence for each candidate. A candidate indicates that the selection criteria are satisfied; subsequent review determines whether the source evidence supports it.
\end{itemize}

\noindent\textbf{Output format.} Each candidate contains the fields below. If no candidate is detected, return an empty list and record the observability status of the cue.
\onecolumn
\par\noindent\begin{minipage}{\linewidth}
\footnotesize
\begin{tabular}{@{}p{0.25\linewidth}p{0.68\linewidth}@{}}
\toprule
Field & Content \\
\midrule
Cue identifier & Corresponding acoustic or visual cue. \\
Candidate identifier & Identifier distinguishing candidates within the same cue. \\
Temporal boundaries & Start and end times in the original media, in seconds. \\
Behavioral measurements & Cue-specific measurement values and units. \\
Quality flags & Information about candidate observation quality. \\
Source references & Feature records and constituent windows supporting the candidate. \\
\bottomrule
\end{tabular}
\end{minipage}\par

\nolinenumbers
\subsection{PHQ-9 Text Analysis and Internal Review}
\label{sec:text-agent-prompts}
\begin{agentprompt}{Box T1: Textual Evidence Extraction Agent.}
\textbf{Role.} You are a textual evidence extractor. Extract evidence related to PHQ-9 symptoms and personal status from the participant's transcript. Do not predict a binary label or estimate questionnaire scores.

\noindent\textbf{Inputs.} The original transcript with segment identifiers and the PHQ-9 symptom definitions.

\noindent\textbf{Task rules.}
\begin{itemize}
\item Read the source and extract symptoms, explicit denials, and personal-status statements expressed by the participant.
\item Preserve the experiencer, time, negation, and contextual qualifications. Distinguish current or temporally unspecified personal experiences from historical or resolved states, other people's experiences, and general discussion.
\item Quotes must come from their cited source segments. Do not join different segments or invent symptoms, frequency, or severity.
\item Each symptom quote must express its assigned item. Ordinary task frustration, boredom with one activity, or general life stress does not automatically establish a symptom.
\item Physical illness, medication, and life events may qualify the evidence or provide alternative explanations, but do not automatically erase an explicitly reported symptom.
\item Record explicit personal depression status and treatment statements separately; they do not replace specific symptom evidence.
\item Absence of mention is not denial. Do not fill unsupported items or count duplicate evidence repeatedly.
\end{itemize}
\noindent\textbf{Output format.} Return structured evidence containing symptom-support records (PHQ-9 item, source segment identifier, quotation, and qualifications), explicit-denial or opposite-experience records (item, source identifier, and quotation), personal-status records (depression or treatment quotation, experiencer, and time), and source-supported alternative explanations and uncertainties.
\end{agentprompt}
\begin{agentprompt}{Box T2: Initial Textual Assessment Agent.}
\textbf{Role.} You are a textual assessment agent. Produce a research binary prediction from structured textual evidence, not a clinical diagnosis.

\noindent\textbf{Inputs.} Extracted symptom support, explicit denials, personal-status statements, alternative explanations, and uncertainties.

\noindent\textbf{Decision rules.}
\begin{itemize}
\item Integrate all evidence. Do not decide solely from a keyword, one symptom, or the number of symptoms.
\item Persistent or explicit loss of interest or low mood, multiple independent current symptoms, functional impairment, and self-harm expressions can strengthen a positive assessment.
\item Reviewed personal current or temporally unspecified depression status and current antidepressant treatment may provide personal-status evidence, but cannot substitute for specific symptoms.
\item Other people's experiences, general discussion, resolved history, and medication of unspecified purpose do not establish the participant's current status.
\item A single mild, temporally unclear, or explicitly explained symptom does not automatically support a positive assessment.
\item Unmentioned symptoms are not counter-evidence. Reflect uncertainty when support and denial coexist.
\item Ground the rationale and key items in the supplied evidence. Do not infer the dataset label or questionnaire score.
\end{itemize}
\noindent\textbf{Output format.} Return a JSON object with the following fields:
\begin{itemize}
\item \texttt{prediction\_label2}: initial label, 0 or 1.
\item \texttt{probability\_label2\_is\_1}: probability of class 1 in $[0,1]$.
\item \texttt{rationale}: principal evidence and reasoning for the assessment.
\item \texttt{key\_item\_numbers}: the PHQ-9 items supporting the assessment.
\item \texttt{uncertainty}: unresolved uncertainty.
\end{itemize}
\end{agentprompt}
\begin{agentprompt}{Box T3: Internal Review Agent.}
\textbf{Role.} You are an independent textual evidence reviewer. Revisit the source to check symptom interpretations and their evidential basis, and produce a review used to determine the final text label T.

\noindent\textbf{Inputs.} Original textual context with segment identifiers and speaker information, without audiovisual predictions or ground-truth labels.

\noindent\textbf{Review rules.}
\begin{itemize}
\item Independently examine the experiencer, symptom meaning, current or historical time, and alternative explanations.
\item Do not interpret depression-related discussion, other people's experiences, education, or resolved history as the participant's current state.
\item Loss of interest requires loss of interest or pleasure. Low mood requires an emotional expression, rather than an inference from physical fatigue or unfinished work.
\item Ordinary failure, polite guilt, everyday forgetfulness, and fatigue after exercise do not automatically establish the corresponding symptoms.
\item Retain genuine personal symptoms even when life stress is present. Consider persistence, functional impact, and independent symptoms together.
\item Interpret indirect self-harm expressions semantically; do not require a particular diagnostic word.
\item Cite only participant statements. Interviewer statements provide context, not participant evidence. Both positive and negative judgments require source citations.
\item Do not infer unstated symptoms, history, or labels.
\end{itemize}
\noindent\textbf{Output format.} Return a JSON object with \texttt{rationale} (a brief review rationale), \texttt{evidence\_ids} (at most three supplied source identifiers), \texttt{prediction\_label2} (0 or 1), and \texttt{probability\_label2\_is\_1} (a value in $[0,1]$ consistent with the label).
\end{agentprompt}
\subsection{Support Verification}
The following templates present the approved acoustic and visual review instructions in English. Output descriptions are translated for presentation.

\begin{agentprompt}{Box A1: Acoustic Support Agent.}
\textbf{Role.} You are an acoustic behavioral evidence reviewer. Determine whether the candidate interval supports the specified acoustic cue using the audio and its corresponding features.

\noindent\textbf{Inputs.} Cue identifier and definition, source audio interval, transcript for locating the utterance, and corresponding acoustic measurements.

\noindent\textbf{Task rules.}
\begin{itemize}
\item Listen to the audio first, then use feature values to check the same acoustic phenomenon. The transcript locates the utterance but does not establish an acoustic observation.
\item Do not invent values, times, or comparison baselines that were not supplied.
\item Describe only what is heard in the auditory observation; place times and numbers in the corresponding feature explanation.
\item Describe the core phenomenon in at most 40 Chinese characters. Cite one or two actual features and explain how they relate to the observation.
\item The verdict must agree with the observation and feature explanation.
\end{itemize}
\noindent\textbf{Cue-specific focus.}
\begin{itemize}
\item A01: Describe loudness variation and cite loudness SD or P90--P10.
\item A02: Describe the rate of pitch changes during continuous speech and cite relative F0 change rate. Small pitch excursions do not establish a low rate of change.
\item A03: Describe articulation speed excluding pauses and cite articulation rate. Slow overall speech caused by frequent pauses does not establish slow articulation.
\item A04: Check whether the pause occurs within the same response and cite its duration.
\item A05: Describe at least two audible pauses and cite their count, events, or durations.
\end{itemize}
\noindent\textbf{Output format.} Return only two lines:
\begin{quote}\small\raggedright
\texttt{VERDICT:} supported or not\_supported\\
\texttt{JOINT\_EXPLANATION:} Heard in the audio: specific acoustic phenomenon; corresponding features: field = actual value, with an explanation of the relationship.
\end{quote}
\end{agentprompt}
\begin{agentprompt}{Box A2: Visual Support Agent.}
\textbf{Role.} You are a visual behavioral evidence reviewer. Determine whether the core interval supports the specified visual cue using consecutive frames and temporally corresponding features.

\noindent\textbf{Inputs.} Cue definition, core interval, consecutive images with identifiers, corresponding frame-level features, and interval statistics.

\noindent\textbf{Task rules.}
\begin{itemize}
\item Review only the core interval. Frame-level features correspond to images with the same identifiers; missing values are not zeros.
\item Observe consecutive frames first, then check the same behavior against temporally corresponding feature values. A positive feature value cannot replace visual observation.
\item Observe at least three adjacent images covering 80\% of the core interval. Context frames cannot replace core observations.
\item Describe only visible actions in the observation; place times and numbers in the feature explanation.
\item Do not invent values, times, or comparison baselines. Do not interpret engineering selection thresholds as clinical thresholds.
\item Describe the core phenomenon in at most 40 Chinese characters. Cite one or two actual features and explain their correspondence with the visual observation.
\item Cite the frame identifier with each frame-level value. Interval statistics may be cited directly as a field and its actual value.
\item The verdict must agree with the explanation. Do not supply evidence absent from the input.
\end{itemize}
\noindent\textbf{Output format.} Return only two lines:
\begin{quote}\small\raggedright
\texttt{VERDICT:} support or not-support\\
\texttt{JOINT\_EXPLANATION:} Seen in the video: specific action; corresponding features: field = actual value, with an explanation of the relationship.
\end{quote}
\end{agentprompt}
\subsection{Counter Challenge}
\begin{agentprompt}{Box A3: Counter Agent.}
\textbf{Role.} You are a behavioral counter-evidence reviewer. Seek specific counter-evidence or confounding evidence that could overturn the core behavioral claim. Do not replace Support with a supporting judgment.

\noindent\textbf{Inputs.} The behavioral claim, core-interval observations, surrounding-context observations, and relevant behavioral measurements and quality information.

\noindent\textbf{Task rules.}
\begin{itemize}
\item Seek specific counter-evidence against the core behavior definition. Context provides interpretation but cannot replace the core state or contribute to the core behavior proportion.
\item Inaudibility does not establish stable loudness or slow pitch changes. A forward-facing head does not establish sustained downward posture.
\item Examine the necessary conditions of the behavior definition. Differences in task, sample, or statistical unit from the literature limit generalization but do not automatically negate an observed behavior.
\item For proportion-based cues, observing an action at least once does not establish a high occurrence proportion.
\item Detection failure, low confidence, duplicate frames, tracking gaps, or camera motion do not establish low activity. Insufficient quality means the claim cannot be verified, not that a behavioral counterexample exists.
\item Use quality information only when it affects the current claim; it cannot replace core-media observation.
\item Missing information is unknown, not normal or abnormal.
\item Failure to find counter-evidence does not support the claim. Leave the validity of counter-evidence and conflicts between observations to the Evidence agent.
\end{itemize}
\noindent\textbf{Judgment categories.} Counter-evidence found: a specific observation may negate the core claim. No counter-evidence found: the claim was checked without finding a specific counterexample; this provides no additional support. Unverifiable: observation or quality is insufficient for reliable verification.

\noindent\textbf{Output format.} Return a JSON object containing only a judgment (counter-evidence found, no counter-evidence found, or unverifiable) and a reason of at most 80 Chinese characters citing core observations and times and explaining the counter-evidence or observation limit. Requests with audiovisual speaker verification enabled additionally return a speaker-verification result: participant, non-participant, or uncertain. Omit this field when that check is disabled.
\end{agentprompt}
\subsection{Evidence Adjudication}
\begin{agentprompt}{Box A4: Evidence Agent.}
\textbf{Role.} You are a behavioral evidence adjudicator. Determine whether a cue is established by the support and counter-evidence records. Do not make a sample-level diagnosis.

\noindent\textbf{Inputs.} Cue definition, Support and Counter records for each interval, source information, and relevant quality limitations.

\noindent\textbf{Adjudication rules.}
\begin{itemize}
\item Each evidence identifier denotes an independent interval. Compare support and counter-evidence separately for that interval.
\item Admit the cue only if at least one interval has a credible core Support observation satisfying the definition that remains valid after counter-evidence review. Block it if all intervals fail to satisfy the definition or cannot be confirmed.
\item A counterexample in one interval does not automatically negate other intervals.
\item Ground reasons in the records. Do not change evidence times, confuse sources, or reinterpret speaker-attribution observations as acoustic-speed judgments.
\item If a condition is not documented, state that it is unspecified and cannot be confirmed; do not claim it is absent.
\item Identify specific contradictory observations and adjudicate within the same core interval. Neighboring context is explanatory only.
\item Counter's failure to find a counterexample adds no support, cannot increase support strength, and cannot compensate for an unsupported Support claim.
\item Examine Counter's observations, reasons, confound explanations, and quality limitations rather than copying its verdict.
\item Insufficient quality limits verification but is not automatically a counterexample. Explicitly check facts such as cross-response pauses or frozen frames that violate necessary conditions. Reading or similar context cannot negate downward posture unless it contradicts the observed posture itself.
\item Execution or formatting failure does not establish counter-evidence. If the raw response is retained, inspect its actual observations without inventing source times.
\end{itemize}
\noindent\textbf{Output format.} Return one JSON object, with the decision first and rationale second:
\begin{quote}\small\raggedright
\texttt{decision}: admit or block\\
\texttt{rationale}: at most 80 Chinese characters stating the specific record-based grounds; identify missing information when evidence is insufficient.
\end{quote}
\end{agentprompt}
\paragraph{Support admission setting.}
Support admission directly uses candidates that pass support review. Evidence admission additionally adjudicates these records using Counter's findings. Both settings use the same downstream behavioral representation and class-comparison method.

\subsection{Conflict-triggered Text Review}
\label{sec:conflict-agent-prompt}
\begin{agentpromptlong}{Box T4: Conflict Review Agent.}
\textbf{Role.} You are a textual evidence reviewer. Examine symptoms and personal-status evidence in the original text and produce a research binary assessment. Conflict review comprises an independent review followed by a final review.

\par\smallskip\noindent\textbf{Step 1: Independent review.}
\textbf{Inputs.} Original textual context and the inventory of retained quotations.

\noindent\textbf{Task rules.}
You are an independent textual evidence reviewer producing a research depression-screening classification, not a clinical diagnosis. You do not know the ground-truth label. Read the source directly, first give a brief rationale checking the experiencer, symptom meaning, current or historical time, and alternative explanations, and then make a judgment.

Class 1 denotes explicit personal current or temporally unspecified depression status, current treatment for depression, or a sufficiently clear combination of depressive symptoms. Class 0 denotes insufficient textual evidence. Do not treat discussion of depression, another person's experiences, psychoeducation, or resolved history as the participant's current state.

Loss of interest requires loss of previous interest or pleasure, not an ordinary dislike. Low mood requires an emotional expression, not physical fatigue, unfinished work, or a product description. Negative self-evaluation is not a failed examination, polite guilt, or quoted reassurance. Sleeping late normally, describing sleep products, mentioning dining locations, everyday forgetfulness, and fatigue after exercise do not automatically constitute symptoms. Distress, trauma, and fear of death in past events do not directly establish current depression or self-harm ideation.

Retain genuine personal symptoms even when life stress is present. Integrate independent symptoms, persistence, functional impairment, and explicit self-harm expressions. Examine indirect self-harm expressions semantically rather than requiring a particular English diagnostic term. Personal medication use and an explicit depression-treatment indication in the same passage may refer to each other; do not infer the purpose of unspecified medication.

Cite only the participant's supplied statements; interviewer statements provide context. Do not infer unstated symptoms, history, or labels. You have not seen any existing classification labels: judge the source independently and do not guess an earlier model conclusion. Negative judgments also require source citations. Do not treat support/counter-evidence item names as semantically verified conclusions.

\noindent\textbf{Output rules.} Output valid JSON only. Write the rationale first, followed by citations, label, probability, and evidence profile. Both positive and negative judgments must cite their source evidence.

\noindent\textbf{Output format.}
\begin{itemize}
\item \texttt{rationale}: a brief review rationale; the prompt specifies at most 120 Chinese characters.
\item \texttt{evidence\_ids}: one to three supplied evidence identifiers.
\item \texttt{prediction\_label2}: 0 or 1.
\item \texttt{probability\_label2\_is\_1}: probability of class 1 in $[0,1]$.
\item \texttt{evidence\_profile}: one of \texttt{explicit\_status}, \texttt{clear\_symptom\_cluster}, \texttt{non\_specific\_or\_alternative}, \texttt{third\_party\_or\_historical}, or \texttt{insufficient\_source}.
\end{itemize}

\par\smallskip\noindent\textbf{Step 2: Final review.}
\textbf{Inputs.} Original textual context, symptom or personal-status evidence to audit, and two anonymized textual judgments A/B, each with a label and rationale.

\noindent\textbf{Task rules.}
You are a text-screening reviewer. Both judgments may be wrong, and their A/B order does not indicate priority. You do not know their authors, the AV label, the ground-truth label, or whether the sample was classified correctly. Make your own final binary judgment directly from the source rather than selecting a winning rationale.

Check whether the statements concern the participant, whether they describe current, historical, or resolved experiences, whether ordinary advice or education has been mistaken for symptoms, and whether genuine symptoms support the overall judgment. Item fields are previous extractions, not established facts. Retaining genuine symptoms and making an overall negative judgment can be consistent; insufficient support is different from explicit counter-evidence.

Assess whether alternative explanations are supported by the source rather than merely possible. Life stress, positive experiences, and other illnesses can coexist with depressive symptoms. Absence of medical consultation or diagnosis does not itself invalidate an established state. Do not use depression-related words, a single frequency, or symptom counts as automatic positive criteria.

The target is text-based screening risk: class 1 denotes explicit personal current or temporally unspecified status, explicit depression treatment, or a sufficiently clear symptom combination. Class 0 denotes source evidence favoring a negative assessment or adequate alternative explanations; it does not require every symptom to be false. Do not equate inability to make a diagnosis with a negative label.

The two rationales are references only. Ground the final conclusion in the source, without favoring a rationale because it is longer, more confident, or lists more items.

\noindent\textbf{Output rules.} Output JSON only. Do not declare an A/B winner, recopy quotations, or add other fields.

\noindent\textbf{Output format.}
\begin{itemize}
\item \texttt{rationale}: one Chinese sentence of at most 160 characters stating the key source basis for the final judgment.
\item \texttt{evidence\_ids}: one to three existing E/S identifiers from the input.
\item \texttt{prediction\_label2}: final label, 0 or 1.
\item \texttt{confidence} ($c$): self-reported probability that the selected label is correct, in $[0.5,1]$; greater uncertainty should be closer to 0.5.
\end{itemize}
\end{agentpromptlong}

\noindent\textbf{External fusion rule.} An initial T--AV conflict triggers this text-only review. The returned \texttt{confidence} is denoted $c$. For a valid review, $c>0.8$ selects the reviewed text label; otherwise the AV label is used. The external selector applies this rule after the review and does not send the AV label to the reviewer.

\linenumbers

\begin{multicols}{2}
\raggedcolumns
\section{Scope and Further Evaluation}
Behavioral cues are nonspecific, and annotations can contain errors. Source links allow inspection but do not establish clinical validity. These cohorts were used during cue and decision development. Using training references alone does not make them untouched prospective evaluations. Recording-level folds do not guarantee speaker separation. Single-run scores do not measure inference variability or isolate the effect of model size. Further work needs external cohorts, repeated inference runs, and policy-specific evidence-quality evaluation. Source recordings and quotations retain their original access restrictions. EDA has not been validated for autonomous clinical use.

\paragraph{Completed validation and further paired diagnostics.}
Section~\ref{sec:human-grounding-protocol} reports the completed human protocol and pooled 91.02\% cue-annotation accuracy. This source-segment check is separate from policy-specific evidence quality. A paired Support--Evidence evaluation would require the same source units, human labels, and downstream settings. It could then measure retained-cue precision and recall, erroneous admission, and missed supported cues. These additional metrics are not reported here. Independently annotated boundaries would also be needed for a free-boundary localization metric.

\paragraph{Evidence quality and assessment utility.}
Evidence quality asks whether a cue is supported by its cited source. Assessment utility asks whether the evidence workflow improves prediction or revision. The completed human check addresses cue agreement on common segments. Admission counts describe selection behavior, not source precision. The final EDA Support model optimizes that selection while retaining source verification. Its classification gains over extended Evidence review do not establish superior annotation quality. Similarly, temporal events preserve local source access without guaranteeing better classification.

Among accepted revisions, the beneficial proportions are 87.50\% on D-Vlog Leave-Query-Out and 58.97\% on LMVD. Harmful changes account for 12.50\% and 41.03\%, respectively. These are conditional on outputs that changed. Their confidence intervals and net accuracy gains are given in Table~\ref{tab:support-revision-summary}. Figure~4 shows completed source-to-assessment examples, while these aggregate revision statistics include successes and failures.

\paragraph{Backbone effects.}
Table~\ref{tab:backbones-complete} reports the backbone comparison. On D-Vlog Leave-Query-Out, 30B-A3B EDA improves accuracy and F1 over direct prediction by 8.81 and 6.39 points. A larger backbone does not always improve EDA. With Evidence admission, 7B exceeds 30B-A3B by 3.12 F1 points. Both evidence organization and backbone choice affect the result.

\section{Extended Related Work}
\label{sec:supp-related-work}

\paragraph{Multimodal and temporal depression modeling.}
Speech, facial activity, and nonverbal behavior provide information for computational depression assessment~\cite{mundt2007voice,girard2014withdrawal,aldeneh2019mood}. Multimodal architectures learn feature interactions and handle unaligned sequences~\cite{zadeh2017tfn,tsai2019mult}. Depression-specific methods use time-aware fusion, instance selection, and multiscale temporal representations~\cite{zhou2023tamfn,zhou2023caiinet,tao2024depmstat,tao2024stst,ling2024mdavif,shangguan2024attentionmil,he2024depressformer}. Recent work also studies state-space models, multiscale interactions, and facial behavior~\cite{ye2024depmamba,zhou2025cmfmamba,zhou2025stcmmamba,shi2026msinet,ma2026bhdd}. EDA focuses on the observations used for assessment. It defines cues, links them to source intervals, and reviews them before comparison.

\paragraph{Symptom evidence and agent deliberation.}
PHQ-9 defines established symptom domains~\cite{kroenke2001phq}. Question-wise Modality Fusion, RED, and Explain-MDRC organize interview content around symptoms and supporting passages~\cite{mandal-etal-2025-enhancing,zhang-etal-2025-explainable,zheng2026explainmdrc}. Dep-LLM and AI Psychiatrist Assistant use evidence and specialized reasoning roles for assessment~\cite{lyu2026depllm,greene2026psychiatrist}. Multimodal language-model studies extend this analysis to speech, vision, and text~\cite{zhao-etal-2026-hears,zhang2026mllmdr}. DepressionAgent is closely related. It combines modality-specific extraction, support--challenge deliberation, arbitration, and reflection~\cite{zhu2026depressionagent}. EDA defines the cue records that agents review and makes their admission into assessment explicit.

\paragraph{Perception, grounding, and source verification.}
VideoAgent, VideoTree, and DrVideo organize long-video content for language-model reasoning~\cite{wang2024videoagent,wang2025videotree,Ma_2025_CVPR}. VideoGLaMM studies fine-grained visual grounding~\cite{munasinghe2025videoglamm}. OPPO distinguishes cue use from input-supported emotion reasoning and uses reinforcement learning to improve perception~\cite{han2026oppo}. EDA performs source checks at inference time with frozen models. Evidence retains cue definitions, source intervals, support, and contextual challenges. Support simplifies admission. Debate and verification methods also examine alternative interpretations~\cite{du2024debate,shi2025aotd}. In EDA, agents review a defined behavioral record. Its source support and later use can be inspected separately.

\FloatBarrier
\end{multicols}
{\small\begin{multicols}{2}\raggedcolumns
\bibliographystyle{ieeenat_fullname}\bibliography{references}
\end{multicols}}
\end{document}